\documentclass[preprint]{elsarticle}
\usepackage{amsmath,amssymb,bm,graphicx,color}
\usepackage{lineno}
\usepackage[colorlinks, linkcolor=red, anchorcolor=blue, citecolor=green]{hyperref}
\usepackage[titletoc]{appendix}
\usepackage{subfigure}
\newcommand{\mbf}[1]{\mathbf{#1}}
\usepackage[table]{xcolor}
\definecolor{mygray}{gray}{.9}
\usepackage{adjustbox}
\usepackage{rotating}

\journal{}

\begin{document}
	
\begin{frontmatter}
		
\title{Modulating Scalable Gaussian Processes for Expressive Statistical Learning}
		
\author[mymainaddress]{Haitao Liu}
\ead{htliu@dlut.edu.cn}

\author[mysecondaryaddress]{Yew-Soon Ong}
\ead{ASYSOng@ntu.edu.sg}

\author[mythirdaddress]{Xiaomo Jiang}
\ead{xiaomojiang2019@dlut.edu.cn}

\author[mymainaddress]{Xiaofang Wang\corref{mycorrespondingauthor}}\cortext[mycorrespondingauthor]{Corresponding author}
\ead{dlwxf@dlut.edu.cn}

\address[mymainaddress]{School of Energy and Power Engineering, Dalian University of Technology, China, 116024}
\address[mysecondaryaddress]{School of Computer Science and Engineering, Nanyang Technological University, Singapore 639798}
\address[mythirdaddress]{Digital Twin Laboratory for Industrial Equipment at Dalian University of Technology, China, 116024.}

\begin{abstract}
For a learning task, Gaussian process (GP) is interested in learning the statistical relationship between inputs and outputs, since it offers not only the prediction mean but also the associated variability. The vanilla GP however struggles to learn complicated distribution with the property of, e.g., heteroscedastic noise, multi-modality and non-stationarity, from massive data due to the Gaussian marginal and the cubic complexity. To this end, this article studies new scalable GP paradigms including the non-stationary heteroscedastic GP, the mixture of GPs and the latent GP, which introduce additional latent variables to modulate the outputs or inputs in order to learn richer, non-Gaussian statistical representation. We further resort to different variational inference strategies to arrive at analytical or tighter evidence lower bounds (ELBOs) of the marginal likelihood for efficient and effective model training. Extensive numerical experiments against state-of-the-art GP and neural network (NN) counterparts on various tasks verify the superiority of these scalable modulated GPs, especially the scalable latent GP, for learning diverse data distributions.
\end{abstract}


\end{frontmatter}

%\linenumbers

\section{Introduction}
Given the input $\mbf{x} \in \mathbb{R}^{d_{\mbf{x}}}$, rather than simply predicting the point estimation of the output $y(\mbf{x})$, we are more interested in inferring the underlying generative process $f:\mathbb{R}^{d_{\mbf{x}}} \mapsto \mathbb{R}$, the distribution of which is most likely to produce the observed data $\mathcal{D}=\{\mbf{X} \in \mathbb{R}^{N \times d_{\mbf{x}}}, \mbf{y} \in \mathbb{R}^{N}\}$, in order to figure out not only the prediction mean but also the associated variability. Along this line, the modeling of the marginal (conditional) distribution $p(\mbf{y}|\mbf{X})$ becomes the central task, which raises from many machine learning scenarios, for example, regression, conditional density estimation~\cite{dutordoir2018gaussian}, data association~\cite{kaiser2019data}, and uncertainty quantification~\cite{bilionis2012multi}.

The well-known Gaussian process (GP)~\cite{williams2006gaussian} is suitable for building our beliefs upon the statistical relationship between the inputs and outputs due to the Bayesian perspective, thus showcasing widespread application in various scenarios~\cite{son2018learning, kandemir2018supervising, li2018hierarchical, li2019gaussian,sun2020multi}. Usually, the GP adopts the Gaussian assumption and the independent and identically distributed (\textit{i.i.d.}) Gaussian noise to conduct efficient closed-form inference and prediction, and employs the stationary kernels $k(.,.)$ to simply quantify how quickly the correlations
vary along each dimension.\footnote{The stationary kernel depends only on the relative distance $||\mbf{x}-\mbf{x}'||$.} Consequently, the vanilla GP may not be appropriate for approximating \textit{non-Gaussuan}, \textit{multi-modal} or \textit{non-stationary} probabilistic behaviors in reality, since its marginal is always Gaussian. Besides, another prominent weakness of  GP is the poor scalability on massive data due to the operations of the $N \times N$ kernel matrix, which raise the cubic complexity $\mathcal{O}(N^3)$. Hence, the above issues raise an urgent demand for having novel GP paradigms which could effectively and efficiently learn rich statistical representations from massive data.

In order to improve the scalability, various scalable GPs have been exploited in recent years from different perspectives~\cite{liu2020gaussian}. Alternatively, we could directly use advanced linear algebra methods, for example, the hierarchical off-diagonal low-rank and matrix vector multiplication algorithms~\cite{ambikasaran2016fast, wang2019exact}, and train the GP distributedly through multiple CPUs/GPUs. Besides, we could resort to the divide-and-conquer idea by splitting the data into small blocks and aggregating predictions from local GPs~\cite{deisenroth2015distributed, liu2018generalized, pillonetto2019distributed}, which naturally support parallel and distributed learning. A more efficient and principled alternative is using sparse approximation~\cite{snelson2006sparse, titsias2009variational, cao2015efficient}, which introduces $M$ ($M \ll N$) global inducing variables $\mbf{u} \in \mathbb{R}^M$ to summarize the latent variables $\mbf{f} = \{\mbf{x}_i\}_{i=1}^N \in \mathbb{R}^N$ statistically, thus significantly reducing the complexity to $\mathcal{O}(M^3)$ through variational inference (VI)~\cite{hensman2013gaussian}. Further reduction of model complexity is possible by exploiting the structured inducing points, see for example~\cite{wilson2015kernel, pleiss2018constant}. This article builds our scalable GPs upon the widely used sparse approximation, which has been further elaborated in Sec.~\ref{sec_sgp}, due to the complete statistical framework and the remarkably low complexity with theoretically guaranteed property~\cite{burt2019rates}.

Aiming to develop new scalable GP paradigms for learning rich probabilistic behaviors, we particularly introduce a Bayesian framework wherein a latent modulation variable $\mbf{w}$ is introduced for encoding the complicated statistical structures from outputs or inputs. Under this modulated GP paradigm, three representative models have been studied, including (i) the scalable heteroscedastic GP (SHGP) which modulates the amplitude of latent output and the noise variance simultaneously, (ii) the scalable mixture of GPs (SMGP) which modulates the assignment of global GP experts at data points for mixing distributions, and finally (iii) the scalable latent GP (SLGP) which augments the input space with latent variables in order to modulate the covariances and moreover the predictive distribution. The key for the above three modulated GPs is to derive scalable and effective evidence lower bounds (ELBOs) for model training.

Along this line, the main contributions of this article are four-fold:
\begin{itemize}
	\item We use the sparse approximation strategy to build a scalable version for SHGP and derive an \textit{analytical} and \textit{scalable} ELBO for efficient training through variational inference;
	\item We sidestep the need of tackling the posterior of discrete assignment distribution for SMGP, and marginalize all the latent variables out to \textit{directly} approximate the marginal likelihood and derive a \textit{tighter} bound for model training with higher quality;
	\item We enhance the power of latent representation of SLGP by introducing a regularized stochastic encoder. Besides, in order to achieve higher training quality, we derive a \textit{hybrid} and \textit{tighter} ELBO that takes the advantage of both the VI and importance-weighted VI (IWVI) based bounds;
	\item We compare the three scalable modulated GPs against state-of-the-art GP and neural network (NN) counterparts to comprehensively investigate their characteristics on various tasks, and release the python implementations at \url{https://github.com/LiuHaiTao01/ModulatedGPs}.
\end{itemize}

The remaining of the article is organized as follows. Sec.~\ref{sec_sgp} first briefly introduces the scalable GP using sparse approximation and variational inference. Thereafter, Sec.~\ref{sec_modulated_gp} develops three scalable GPs that modulate the probabilistic behavior from outputs or inputs, and derives the analytical or tight ELBOs for model training. Sec.~\ref{sec_related_work} then discussed the related works and their differences to our work. Finally, Sec.~\ref{sec_exp} conducts extensive experiments to showcase the superiority of modulated GPs on various tasks, followed by the overall concluding remarks summarized in Sec.~\ref{sec_conclusion}. Note that for improving the readability of this article, the employed acronyms and notations are summarized in Appendix~\ref{app_notation}.

\section{Scalable GP revisited} \label{sec_sgp}
The GP learns the underlying, noise-free latent function $f: \mathbb{R}^{d_{\mbf{x}}} \mapsto \mathbb{R}$ by placing the GP prior over the functional space as $f(\mbf{x}) \sim \mathcal{GP}(0, k(\mbf{x},\mbf{x}'))$, 
where we take the zero mean without loss of generality, and $k(.,.)$ is the kernel function describing the smoothness of $f$.\footnote{The squared exponential (SE) kernel with automatic relevance determination $k(\mbf{x},\mbf{x}') = \nu_f \exp(-0.5 (\mbf{x}-\mbf{x}')^{\mathsf{T}} \bm{\Delta}^{-1} (\mbf{x}-\mbf{x}'))$ is commonly used in practice.} The final observation polluted with independent and identically distributed (\textit{i.i.d.}) Gaussian noise is expressed as
\begin{align} \label{eq_gp}
y(\mbf{x}) = f(\mbf{x}) + \epsilon,
\end{align}
where the observation noise $\epsilon \sim \mathcal{N}(0, \nu^{\epsilon})$. Given $N$ training data $\mathcal{D} = \{\mbf{X}, \mbf{y} \}$, in order to infer the model hyperparameters, we marginalize all the latent variables out and maximize the marginal likelihood
\begin{align}
p(\mbf{y}) = \int p(\mbf{y}|\mbf{f}) p(\mbf{f}) d\mbf{f} = \mathcal{N}(\mbf{y}|\mbf{0}, \mbf{K}_{NN} + \nu^{\epsilon} \mbf{I}),
\end{align}
where the GP prior $p(\mbf{f}) = \mathcal{N}(\mbf{f}|\mbf{0}, \mbf{K}_{NN})$ with the covariance matrix $\mbf{K}_{NN} = k(\mbf{X}, \mbf{X}) \in \mathbb{R}^{N \times N}$, and the Gaussian likelihood $p(\mbf{y}|\mbf{f}) = \mathcal{N}(\mbf{y}|\mbf{f}, \nu^{\epsilon}\mbf{I})$.\footnote{We omit the dependency of these distributions on the deterministic inputs $\mbf{X}$.} After model training, we calculate the Bayes' rule $p(\mbf{f}|\mbf{y}) \propto p(\mbf{y}|\mbf{f}) p(\mbf{f})$ and use it to predict at the test point $\mbf{x}_*$ as $p(f_*|\mbf{y}) = \int p(f_*|\mbf{f}) p(\mbf{f}|\mbf{y}) d\mbf{f} = \mathcal{N}(f_*|\mu_*^f, \nu_*^f)$ where the prediction mean $\mu_*^f = \mbf{k}_{*N}(\mbf{K}_{NN} + \nu^{\epsilon} \mbf{I})^{-1} \mbf{y}$ and the prediction variance $\nu_*^f = k_{**} - \mbf{k}_{*N}(\mbf{K}_{NN} + \nu^{\epsilon} \mbf{I})^{-1}\mbf{k}_{N*}$ with $k_{**} = k(\mbf{x}_*, \mbf{x}_*) \in \mathbb{R}$ and $\mbf{k}_{*N} = \mbf{k}_{N*}^{\mathsf{T}} = k(\mbf{x}_*, \mbf{X}) \in \mathbb{R}^{1 \times N}$. Furthermore, we have the final predictive distribution $p(y_*|\mbf{y}) = \int p(y_*|f_*) p(f_*|\mbf{y})df_* = \mathcal{N}(y_*|\mu_*^f, \nu_*^f + \nu^{\epsilon})$. 

The vanilla GP however suffers from poor scalability due to the cubic complexity $\mathcal{O}(N^3)$ raised by the operations over the $N \times N$ kernel matrix $\mbf{K}_{NN}$. To alleviate this issue in the era of big data, the well-known scalable GP (SGP) using sparse approximation~\cite{snelson2006sparse, titsias2009variational} introduces $M$ ($M \ll N$) inducing variables $\mbf{u}$ at the inducing points $\mbf{Z} \in \mathbb{R}^{M \times d_{\mbf{x}}}$ as sufficient statistics of $\mbf{f}$. Consequently, we arrive at the Nystr\"om approximation $\mbf{K}_{NN} \approx \mbf{K}_{NM} \mbf{K}_{MM}^{-1} \mbf{K}_{MN}$, where $\mbf{K}_{NM} = k(\mbf{X},\mbf{Z}) \in \mathbb{R}^{N \times M}$ and $\mbf{K}_{MM} = k(\mbf{Z},\mbf{Z}) \in \mathbb{R}^{M \times M}$, with the time complexity reduced as $\mathcal{O}(NM^2)$. 

Furthermore, the variational inference employs a tractable variational posterior $q(\mbf{u}) = \mathcal{N}(\mbf{u}|\mbf{m},\mbf{S})$ to minimize the Kullback-Leibler (KL) divergence
\begin{align*}
\mathrm{KL}[q(\mbf{f}, \mbf{u})||p(\mbf{f}, \mbf{u}|\mbf{y})] = \log p(\mbf{y}) - \mathcal{L}_{\mathrm{sgp}},
\end{align*}
which is equivalent to maximizing the analytical ELBO $\mathcal{L}_{\mathrm{sgp}}$ expressed as~\cite{hensman2013gaussian}
\begin{align} \label{eq_elbo_svgp}
\begin{split}
\mathcal{L}_{\mathrm{sgp}} =& \mathbb{E}_{q(\mbf{f})} [\log p(\mbf{y}|\mbf{f})] - \mathrm{KL}(q(\mbf{u})||p(\mbf{u})) \\
=& \sum_{i=1}^N \mathbb{E}_{q(f_i)} [\log p(y_i|f_i)] - \mathrm{KL}(q(\mbf{u})||p(\mbf{u})),
\end{split}
\end{align}
where the GP prior $p(\mbf{u}) = \mathcal{N}(\mbf{u}|\mbf{0},\mbf{K}_{MM})$, the posterior $q(\mbf{f}) = \int p(\mbf{f}|\mbf{u}) q(\mbf{u}) d\mbf{u} =\mathcal{N}(\mbf{f}|\bm{\mu}, \bm{\Sigma})$ with the mean and covariance expressed respectively as
\begin{align*}
\bm{\mu}^f =& \mbf{K}_{NM}\mbf{K}_{MM}^{-1}\mbf{m}, \\
\bm{\Sigma}^f =& \mbf{K}_{NN}-\mbf{K}_{NM}\mbf{K}_{MM}^{-1}[\mbf{I}-\mbf{S}\mbf{K}_{MM}^{-1}]\mbf{K}_{MN},
\end{align*}
and the individual $q(f_i) = \mathcal{N}(f_i|\mu^f_i\triangleq[\bm{\mu}^f]_i, \nu^f_i\triangleq[\bm{\Sigma}^f]_{ii})$. Due to the factorization of the expectation term in~\eqref{eq_elbo_svgp} over data points, the bound has an unbiased estimation
\begin{align}
\mathcal{L}_{\mathrm{sgp}} \approx \frac{N}{|\mathcal{B}|} \sum_{i \in \mathcal{B}} \mathbb{E}_{q(f_i)} [\log p(y_i|f_i)] - \mathrm{KL}(q(\mbf{u})||p(\mbf{u})),
\end{align}
where $\mathcal{B}$ is a subset of the training data $\mathcal{D}$. This bound owns a remarkable complexity of $\mathcal{O}(M^3)$ through the stochastic optimization with single-sample approximation (i.e., $|\mathcal{B}|=1$).

It is observed that the vanilla SGP only (i) provide the Gaussian predictive distribution with homoscedastic noise; and (ii) describe the invariant spatio-temporal behaviors due to the stationary kernels, which therefore limit the application to complicated tasks dominated by rich underlying stochastic processes.

\section{Modulated scalable Gaussian processes} \label{sec_modulated_gp}
It is known that the vanilla GP paradigm struggles to approximate complicated distribution. Hence, in order to improve the capability, we introduce an additional latent variable $\mbf{w}$ to modulate the behavior of GP as 
\begin{align}
y(\mbf{x}) = f(\mbf{x}, \mbf{w}).
\end{align}
The latent variable $\mbf{w}$ permits disentangling the statistical structure to enhance the expressivity of modeling. Hence, the challenging details in data, e.g., the heteroscedastic noise, the multi-modality, and the non-stationarity, could be explained and absorbed by $\mbf{w}$. We next proceed to present three scalable modulated GPs which perform the modulation on the outputs or inputs.

\begin{figure}[t!]
	\centering
	\includegraphics[width=1.\textwidth]{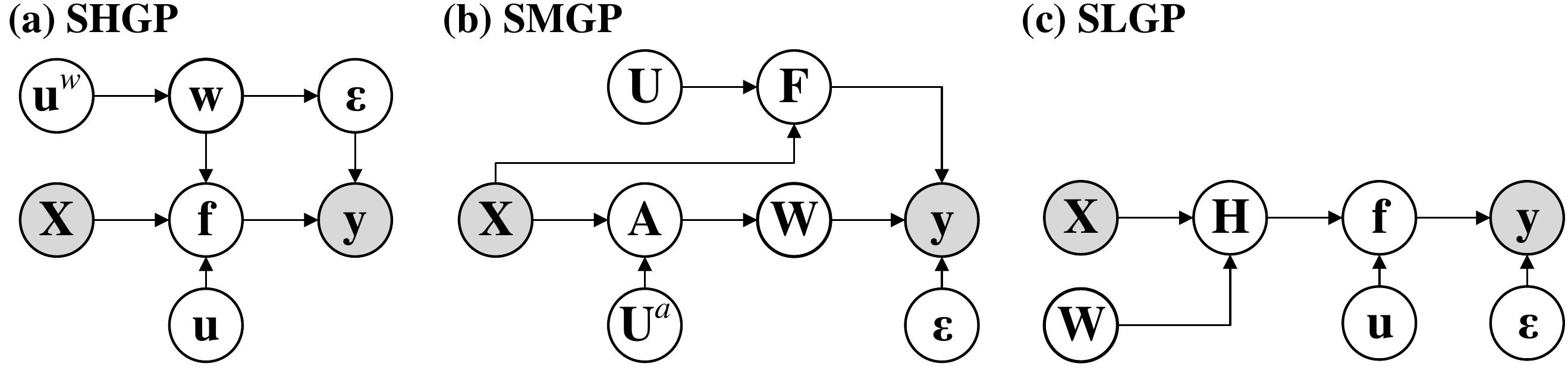}
	\caption{The graphical models of three scalable modulated GPs, wherein the (a) SHGP and (b) SMGP perform the modulation on the output (noise), while the (c) SLGP modulates the inputs.}
	\label{fig_modulated_gps}
\end{figure}

\subsection{Scalable heteroscedastic GP}
Instead of using the poor \textit{i.i.d} noise, the heteroscedastic GP (HGP) defines the following additive model
\begin{align} \label{eq_hgp}
y(\mbf{x}) = e^{w(\mbf{x})} f(\mbf{x}) + \epsilon(\mbf{x})
\end{align}
to capture the variability in both outputs and noise. In~\eqref{eq_hgp}, the additional latent function $w(.): \mathbb{R}^{d_{\mbf{x}}} \mapsto \mathbb{R}$ is introduced to modulate the amplitude of output $f(.)$ in order to describe the non-stationarity~\cite{adams2008gaussian}; besides, it has an input-dependent noise $\epsilon(\mbf{x}) = \mathcal{N}(0, c e^{2w(\mbf{x})})$~\cite{goldberg1998regression}, which also uses $w(.)$ to modulated the noise variance; and the positive parameter $c$ leaves flexibility for adjusting the noise variance. Note that the exponential form of $w(.)$ ensures the positivity of noise variance and the modulation for $f$. 

The HGP takes the following GP priors
\begin{align}
p(\mbf{f}) = \mathcal{N}(\mbf{f}|\mbf{0}, \mbf{K}_{NN}), \quad p(\mbf{w}) = \mathcal{N}(\mbf{w}|\mu_0^w \mbf{1}, \mbf{K}_{NN}^w),
\end{align}
and the \textit{non-stationary}, \textit{heteroscedastic} likelihood
\begin{align}
p(\mbf{y}|\mbf{f}, \mbf{w}) = \mathcal{N}(\mbf{y}|e^{\mbf{w}}\mbf{f}, \mbf{\Sigma}^{\epsilon}),
\end{align}
where the covariance $\mbf{\Sigma}^{\epsilon} = \mathrm{diag}[ce^{2\mbf{w}}]$. Note that the prior mean $\mu_0^w$ is utilized to account for the variability of noise variance. Besides, it has been pointed out that the likelihood of this model is \textit{log-concave} on both $f$ and $w$, thus resulting in unimodal posteriors~\cite{ munoz-gonzalez2014divisive, munoz-gonzalez2016laplace}.

\textbf{Analytical ELBO.} We next proceed to use variational inference to derive the \textit{scalable} and \textit{analytical} objective for model training of scalable HGP (SHGP), the graphical model of which is depicted in Fig.~\ref{fig_modulated_gps}(a). We first introduce the inducing variables $\mbf{u}$ and $\mbf{u}^w \sim \mathcal{N}(\mbf{m}^w, \mbf{S}^w)$ for $\mbf{f}$ and $\mbf{w}$ respectively. Thereafter, by minimizing the KL divergence $\mathrm{KL}[q(\mbf{f},\mbf{w},\mbf{u},\mbf{u}^w) || p(\mbf{f},\mbf{w},\mbf{u},\mbf{u}^w|\mbf{y})]$ where $q(\mbf{f},\mbf{w},\mbf{u},\mbf{u}^w) = p(\mbf{f}|\mbf{u}) p(\mbf{w}|\mbf{u}^w) q(\mbf{u}) q(\mbf{u}^w)$, we arrive at the following ELBO for $\log p(\mbf{y})$ as
\begin{align} \label{eq_elbo_hgp}
\mathcal{L}_{\mathrm{shgp}} = \mathbb{E}_{q(\mbf{w}) q(\mbf{f})} [\log p(\mbf{y}|\mbf{f}, \mbf{w})] - \mathrm{KL}[q(\mbf{u})||p(\mbf{u})] - \mathrm{KL}[q(\mbf{u}^w)||p(\mbf{u}^w)],
\end{align}
where the GP prior $p(\mbf{u}^w) = \mathcal{N}(\mbf{u}^w|\mbf{0},\mbf{K}^w_{MM})$, and the variational posterior $q(\mbf{w}) = \int p(\mbf{w}|\mbf{u}^w) q(\mbf{u}^w) d\mbf{u}^w = \mathcal{N}(\mbf{w}|\bm{\mu}^w, \bm{\Sigma}^w)$ with the mean $\bm{\mu}^w = \mbf{K}_{NM}^w [\mbf{K}^w_{MM}]^{-1}(\mbf{m}^w - \mu_0^w \mbf{1}) + \mu_0^w \mbf{1}$ and covariance $\bm{\Sigma}^w = \mbf{K}^w_{NN}-\mbf{K}^w_{NM}[\mbf{K}^w_{MM}]^{-1}[\mbf{I}-\mbf{S}^w[\mbf{K}^w_{MM}]^{-1}]\mbf{K}^w_{MN}$. 
Due to the Gaussian posteriors $q(\mbf{u})$ and $q(\mbf{u}^w)$, the two KL terms in~\eqref{eq_elbo_hgp} have closed-form expressions. Furthermore, the expectation of the likelihood in the right-hand side of $\mathcal{L}_{\mathrm{shgp}}$ can be analytically expressed as
\begin{align}
\begin{split}
&\mathbb{E}_{q(\mbf{w}) q(\mbf{f})} [\log p(\mbf{y}|\mbf{f}, \mbf{w})] \\
=& \mathbb{E}_{q(\mbf{w})} \left[\log \mathcal{N}(\mbf{y}|\bm{\mu}^f, \mbf{\Sigma}^{\epsilon}) - \frac{1}{2}\mathrm{Tr}[(\mbf{\Sigma}^{\epsilon})^{-1} \bm{\Sigma}^f ] \right] \\
=& -\frac{n}{2} \log (2c\pi) - \frac{1}{2} \sum_{i=1}^n \left( 2 [\bm{\mu}^w]_i + \frac{1}{c}([\mbf{R}_1^w]_{ii} y_i^2 - 2y_i[\mbf{R}_2^{w}\bm{\mu}^f]_i + [\bm{\mu}^f]_i^2 +[\mbf{\Sigma}^f]_{ii}) \right),
\end{split}
\end{align}
where the diagonal matrices $\mbf{R}_1^w = \mathrm{diag}[e^{2\bm{\nu}^w-2\bm{\mu}^w}]$ and $\mbf{R}_2^w = \mathrm{diag}[e^{0.5 \bm{\nu}^w - \bm{\mu}^w}]$ with $\bm{\nu}^w$ being the vector comprising the diagonal elements of $\bm{\Sigma}^w$.

\textbf{Prediction.} Finally, when performing prediction at the test point $\mbf{x}_*$, we have 
\begin{align}
\begin{split}
p(y_*|\mbf{y}) =& \int p(y_*|f_*,w_*) p(f_*|\mbf{y}) p(w_*|\mbf{y}) df_* dw_* \\
=& \int \mathcal{N}(y_*|\mu^f_* e^{w_*}, e^{2w_*}\nu^f_* + ce^{2w_*}) p(w_*|\mbf{y}) dw_*,
\end{split}
\end{align}
where the prediction $p(f_*|\mbf{y}) = \int p(f_*|\mbf{u}) q(\mbf{u}) d\mbf{u} = \mathcal{N}(f_*|\mu_*^f, \nu_*^f)$ with the mean $\mu_*^f = \mbf{k}_{*M}\mbf{K}_{MM}^{-1}\mbf{m}$ and variance $\nu_*^f = k_{**}-\mbf{k}_{*M}\mbf{K}_{MM}^{-1}[\mbf{I}-\mbf{S}\mbf{K}_{MM}^{-1}]\mbf{k}_{M*}$; and similarly, $p(w_*|\mbf{y}) = \int p(w_*|\mbf{u}^w) q(\mbf{u}^w) d\mbf{u}^w = \mathcal{N}(w_*|\mu_*^w, \nu_*^w)$ with the mean $\mu_*^w = \mbf{k}^w_{*M}[\mbf{K}^w_{MM}]^{-1}(\mbf{m}^w - \nu_0^w\mbf{1}) + \nu_0^w\mbf{1}$ and variance $\nu_*^w = k^w_{**}-\mbf{k}^w_{*M}[\mbf{K}^w_{MM}]^{-1}[\mbf{I}-\mbf{S}^w[\mbf{K}^w_{MM}]^{-1}]\mbf{k}^w_{M*}$.
The prediction $p(y_*|\mbf{y})$ can be estimated through the Markov Chain Monte Carlo (MCMC) sampling, or more efficiently, the Gauss-Hermite quadrature. Note that the predictive distribution of SHGP is \textit{non-Gaussian}.

\subsection{Scalable mixture of GPs} \label{sec_mgp}
The capability of SHGP for describing complicated distribution is limited due to the Gaussian noise. However, it is known that the mixture of infinite Gaussians could approximate any target distribution. To this end, the mixture of GPs (MGP) aggregates the predictions from GP experts in order to break through the Gaussian assumption. By activating the GP experts in different locations, the MGP is enabled to tackle both the \textit{non-stationary} and \textit{multi-modal} data distributions.

Specifically, the MGP employs $T$ independent latent functions $\{f^t\}_{t=1}^T$, and the indicator vector $\mbf{w}_i \in \{0,1\}^T$ at point $\mbf{x}_i$ to represent its assignment to one of the GP experts. Consequently, we have the following likelihood
\begin{align} \label{eq_mgp}
p(\mbf{y}|\mbf{F},\mbf{W}) = \prod_{i=1}^N \prod_{t=1}^T p(y_i|f_i^t, \nu^t)^{\mathbb{I}(w_i^t=1)},
\end{align}
where the sets $\mbf{F} = \{\mbf{f}_i \in \mathbb{R}^T\}_{i=1}^{N} = \{\mbf{f}^t \in \mathbb{R}^N \}_{t=1}^T$, $\mbf{W} = \{\mbf{w}_i \in \mathbb{R}^T\}_{i=1}^{N} = \{\mbf{w}^t \in \mathbb{R}^N \}_{t=1}^T$, the variable $\nu^t$ is the noise variance for the $t$-th GP expert, and $\mathbb{I}(.)$ is the indicator function. It is observed that~\eqref{eq_mgp} defines a generative process where the observed data is generated by the mixture of several global independent stochastic processes.

We further assume that the indicator $\mbf{w}_i$ are drawn independently from a multinomial distribution with unnormalized logit parameters $\mbf{a}_i = \{a_i^t = a^t(\mbf{x}_i)\}_{t=1}^T$ as
\begin{align}
p(\mbf{w}_i|\mbf{a}_i) = \mathcal{M}(\mbf{w}_i|\mathtt{Softmax}(\mbf{a}_i)), \quad 1\le i \le N,
\end{align}
where the logit parameters are assumed to follow independent GP priors as
\begin{align}
p(\mbf{a}^t) = \mathcal{N}(\mbf{a}^t|\mbf{0}, \mbf{K}^{a,t}_{NN}), \quad 1 \le t \le T.
\end{align}
Now the GP priors $p(\mbf{f}^t) = \mathcal{N}(\mbf{f}^t|\mbf{0}, \mbf{K}^{t}_{NN})$ and $p(\mbf{a}^t)$ in MGP encode the structures of both functions and associations.

\textbf{Tighter ELBO.} We thereafter decide to build the scalable training framework for the scalable MGP (SMGP) depicted in Fig.~\ref{fig_modulated_gps}(b). The joint prior of SMGP is defined as
\begin{align*}
p(\mbf{y},\mbf{F},\mbf{U},\mbf{W},\mbf{A} ,\mbf{U}^a) = p(\mbf{y}|\mbf{F}, \mbf{W}) p(\mbf{F}|\mbf{U}) p(\mbf{U}) p(\mbf{W}|\mbf{A}) p(\mbf{A}|\mbf{U}^a) p(\mbf{U}^a).
\end{align*}
where the sets $\mbf{A} = \{\mbf{a}_i \in \mathbb{R}^T \}_{i=1}^N = \{\mbf{a}^t \in \mathbb{R}^N \}_{t=1}^T$ and $\mbf{U}^a = \{\mbf{u}^{a,t} \in \mathbb{R}^M \}_{t=1}^T$ are the inducing variables for $T$ assignment GPs;\footnote{For the convenience of presentation, we assume that the inducing sizes for the GPs are the same.} the conditionals factorize as $p(\mbf{F}|\mbf{U}) = \prod_{t=1}^T p(\mbf{f}^t|\mbf{u}^t)$ and $p(\mbf{W}|\mbf{A}) = \prod_{i=1}^N p(\mbf{w}_i|\mbf{a}_i)$; and finally, the GP priors $p(\mbf{U}) = \prod_{t=1}^T p(\mbf{u}^{t}) = \prod_{t=1}^T \mathcal{N}(\mbf{u}^{t}|\mbf{0}, \mbf{K}_{MM}^{t})$ and $p(\mbf{U}^a) = \prod_{t=1}^T p(\mbf{u}^{a,t}) = \prod_{t=1}^T \mathcal{N}(\mbf{u}^{a,t}|\mbf{0}, \mbf{K}_{MM}^{a,t})$. Correspondingly, the joint variational posterior writes as
\begin{align*}
q(\mbf{F},\mbf{U},\mbf{W},\mbf{A} ,\mbf{U}^a) = p(\mbf{F}|\mbf{U}) q(\mbf{U}) q(\mbf{W}|\mbf{A}) p(\mbf{A}|\mbf{U}^a) q(\mbf{U}^a),
\end{align*}
where the variational posteriors $q(\mbf{U}) = \prod_{t=1}^T q(\mbf{u}^{t}) = \prod_{t=1}^T \mathcal{N}(\mbf{u}^{t}|\mbf{m}^t, \mbf{S}^{t})$ and $q(\mbf{U}^a) = \prod_{t=1}^T q(\mbf{u}^{a,t}) = \prod_{t=1}^T \mathcal{N}(\mbf{u}^{a,t}|\mbf{m}^{a,t}, \mbf{S}^{a,t})$.

The variational inference then helps derive the following ELBO for SMGP as
\begin{align} \label{eq_elbo_vi_mgp}
\begin{split}
\mathcal{L} =& \mathbb{E}_{q(\mbf{F})q(\mbf{W}|\mbf{A})q(\mbf{A})} [\log p(\mbf{y}|\mbf{F}, \mbf{W})] - \mathrm{KL}[q(\mbf{W}|\mbf{A})||p(\mbf{W}|\mbf{A}))] \\
&- \mathrm{KL}[q(\mbf{U})||p(\mbf{U})] - \mathrm{KL}[q(\mbf{U}^a)||p(\mbf{U}^a)],
\end{split}
\end{align}
where the posterior factorizes as $q(\mbf{A}) = \prod_{t=1}^T \int p(\mbf{a}^t|\mbf{u}^{a,t}) q(\mbf{u}^{a,t}) d\mbf{u}^{a,t} = \prod_{t=1}^T \mathcal{N}(\mbf{a}^t|\bm{\mu}^{a,t}, \bm{\Sigma}^{a,t})$ with the mean $\bm{\mu}^{a,t} = \mbf{K}_{NM}^{a,t} [\mbf{K}^{a,t}_{MM}]^{-1}\mbf{m}^{a,t}$ and covariance $\bm{\Sigma}^{a,t} = \mbf{K}^{a,t}_{NN}-\mbf{K}^{a,t}_{NM}[\mbf{K}^{a,t}_{MM}]^{-1}[\mbf{I}-\mbf{S}^{a,t}[\mbf{K}^{a,t}_{MM}]^{-1}]\mbf{K}^{a,t}_{MN}$; and similarly, the variational posterior $q(\mbf{F}) = \prod_{t=1}^T p(\mbf{f}^t|\mbf{u}^t) q(\mbf{u}^t) d\mbf{u}^t= \prod_{t=1}^T \mathcal{N}(\mbf{f}^t|\bm{\mu}^{f,t}, \bm{\Sigma}^{f,t})$ with the mean $\bm{\mu}^{f,t} = \mbf{K}_{NM}^{t} [\mbf{K}^{t}_{MM}]^{-1}\mbf{m}^{t}$ and covariance $\bm{\Sigma}^{f,t} = \mbf{K}^{t}_{NN}-\mbf{K}^{t}_{NM}[\mbf{K}^{t}_{MM}]^{-1}[\mbf{I}-\mbf{S}^{t}[\mbf{K}^{t}_{MM}]^{-1}]\mbf{K}^{t}_{MN}$. However, since the prior $p(\mbf{W}|\mbf{A}) = \prod_{i=1}^N p(\mbf{w}_i|\mbf{a}_i)$ is the product of discrete multinomial distributions, it is not straightforward to handle the posterior $q(\mbf{W}|\mbf{A})$ and the related KL divergence in~\eqref{eq_elbo_vi_mgp}. To sidestep this issue, Kaiser et al.~\cite{kaiser2019data} decided to keep the latent variables $\mbf{W}$ by approximating $\log p(\mbf{y}, \mbf{W})$ rather than the interested $\log p(\mbf{y})$.

Differently, we here marginalize all the latent variables out to \textit{directly} approximate $\log p(\mbf{y})$ through a \textit{tighter} ELBO. To this end, we first follow the sparse GP framework to derive the lower bound for the conditional $\log p(\mbf{y}|\mbf{W})$ as
\begin{align}
\begin{split}
\mathcal{L}_{\mbf{W}} =& \mathbb{E}_{q(\mbf{F})} [\log p(\mbf{y}|\mbf{F}, \mbf{W})] - \mathrm{KL}[q(\mbf{U})||p(\mbf{U})] \\
=& \tilde{\mathcal{L}}_{\mbf{W}}^{\mathrm{mgp}} - \mathrm{KL}[q(\mbf{U})||p(\mbf{U})].
\end{split}
\end{align} 
Thereafter, according to Jensen's inequality we have
\begin{align} \label{eq_elbo_mgp}
\begin{split}
\log p(\mbf{y}) \ge& \log \int \exp(\tilde{\mathcal{L}}_{\mbf{W}}^{\mathrm{mgp}}) p(\mbf{W}|\mbf{A}) p(\mbf{A}|\mbf{U}^a) \frac{p(\mbf{U}^a)}{q(\mbf{U}^a)} q(\mbf{U}^a) d\mbf{W} d\mbf{A} d\mbf{U}^a \\
\ge& \mathbb{E}_{ q(\mbf{A})} \left[\log \mathbb{E}_{p(\mbf{W}|\mbf{A})}[\exp(\tilde{\mathcal{L}}_{\mbf{W}}^{\mathrm{mgp}})]  \right] - \mathrm{KL}[q(\mbf{U})||p(\mbf{U})] - \mathrm{KL}[q(\mbf{U}^a)||p(\mbf{U}^a)] \\
=& \mathcal{L}_{\mathrm{smgp}} \ge \mathcal{L}.
\end{split} 
\end{align}
It is observed that in comparison to the bound~\eqref{eq_elbo_vi_mgp}, the tighter bound $\mathcal{L}_{\mathrm{smgp}}$ avoids introducing an additional variational posterior $q(\mbf{W}|\mbf{A})$.
Besides, note that during the model training, instead of sampling from the discrete prior distribution $\mbf{w}_i \sim \mathcal{M}(\mbf{w}_i|\mathtt{Softmax}(\mbf{a}_i))$, we employ the continuous relaxation proposed in~\cite{maddison2017the, jang2017categorical} in order to perform back propagation for stochastic optimization: it adopts the reparameterization trick to sample from a Concrete random variable controlled by a temperature parameter $\lambda$, and approaches a discrete random variable when $\lambda \to 0$. Our SMGP adopts a small value of $\lambda = 0.01$.

\textbf{Prediction.} When performing prediction at the test point $\mbf{x}_*$, we have
\begin{align*}
p(y_*|\mbf{y}) = \int p(y_*|\mbf{f}_*, \mbf{w}_*)  p(\mbf{f}_*|\mbf{y}) p(\mbf{w}_*|\mbf{y}) d\mbf{w}_* d\mbf{f}_*,
\end{align*}
where the prediction $p(\mbf{f}_*|\mbf{y}) = \prod_{t=1}^T \int p(f^t_*|\mbf{u}^t) q(\mbf{u}^t) d\mbf{u}^t = \prod_{t=1}^T \mathcal{N}(f^t_*|\mu^{f,t}_*, \nu^{f,t}_*)$ with the mean $\mu_*^{f,t} = \mbf{k}^t_{*M}[\mbf{K}^t_{MM}]^{-1}\mbf{m}^t$ and variance $\nu_*^{f,t} = k^t_{**}-\mbf{k}^t_{*M}[\mbf{K}^t_{MM}]^{-1}[\mbf{I}-\mbf{S}^t[\mbf{K}^t_{MM}]^{-1}]\mbf{k}^t_{M*}$; and the assignment prediction $p(\mbf{w}_*|\mbf{y}) = \int p(\mbf{w}_*|\mbf{a}_*) p(\mbf{a}_*|\mbf{y}) d\mbf{a}_* $ where $p(\mbf{a}_*|\mbf{y}) = \prod_{t=1}^T \int p(a^t_*|\mbf{u}^{a,t}) q(\mbf{u}^{a,t}) d\mbf{u}^{a,t} = \mathcal{N}(a^t_*|\mu^{a,t}_*, \nu^{a,t}_*)$ with the mean $\mu_*^{a,t} = \mbf{k}^{a,t}_{*M}[\mbf{K}^{a,t}_{MM}]^{-1}\mbf{m}^{a,t}$ and variance $\nu_*^{a,t} = k^{a,t}_{**}-\mbf{k}^{a,t}_{*M}[\mbf{K}^{a,t}_{MM}]^{-1}[\mbf{I}-\mbf{S}^{a,t}[\mbf{K}^{a,t}_{MM}]^{-1}]\mbf{k}^{a,t}_{M*}$. Thereafter, we sample from the Gaussian and multinomial distributions, and pass through the model to produce predictive samples.

\subsection{Scalable latent GP}
The foregoing models including SHGP and SMGP directly modulate the outputs for capturing complicated distributions. Differently, we could introduce the stochasticity in the augmented input space, and modulate the covariances in order to output expressive predictive distribution more flexibly. It has been found that the GP could warp the normal stochasticity in the augmented inputs into complicated distribution like~\cite{damianou2014variational,lawrence2005probabilistic}. Along this line, the latent GP (LGP) introduces additional latent inputs $\mbf{w} \in R^{d_{\mbf{w}}}$ to augment the input space as
\begin{align}
y(\mbf{x}) = f([\mbf{x}, \mbf{w}]) + \epsilon.
\end{align}
It is usually assumed that the latent variables $\mbf{W} = \{\mbf{w}_i \}_{i=1}^N$ for data points are independent from each other. We next attempt to derive scalable ELBO for LGP using variational inference.

\textbf{The IWVI-based ELBO.} Similar to SMGP, the scalable LGP (SLGP) first obtains the bound for the conditional $\log p(\mbf{y}|\mbf{W})$ as 
\begin{align} \label{eq_logp(y|w)_lgp}
\log p(\mbf{y}|\mbf{W}) \ge \tilde{\mathcal{L}}_{\mbf{W}}^{\mathrm{lgp}} - \mathrm{KL}[q(\mbf{u})||p(\mbf{u})] = \log \hat{p}(\mbf{y}|\mbf{W}),
\end{align}
where the partial bound $\tilde{\mathcal{L}}_{\mbf{W}}^{\mathrm{lgp}} = \mathbb{E}_{q(\mbf{f}|\mbf{W})} [\log p(\mbf{y}|\mbf{f})]$ is analytical; the posterior $q(\mbf{f}|\mbf{W}) = \int p(\mbf{f}|\mbf{u}, \mbf{W}) q(\mbf{u}) d\mbf{u}= \mathcal{N}(\mbf{f}|\bm{\mu}^f, \bm{\Sigma}^f)$ wherein the kernel matrices in $\bm{\nu}^f$ and $\bm{\Sigma}^f$ are calculated in the augmented input space $\mathbb{R}^{d_{\mbf{x}}+d_{\mbf{w}}}$; and finally, $\hat{p}(\mbf{y}|\mbf{W})$ is an unnormalized distribution which satisfies $ \lim_{\mbf{u} \to \mbf{f}} \log \hat{p}(\mbf{y}|\mbf{W}) = \log p(\mbf{y}|\mbf{W})$,
since in this case the sparse GP recovers the full GP. Thereafter, by inserting the inequality~\eqref{eq_logp(y|w)_lgp} back into $\log p(\mbf{y})$ and using the Jensen's inequality, we get a scalable bound for $\log p(\mbf{y})$ as
\begin{align} \label{eq_elbo_lgp_tight}
\begin{split}
\log p(\mbf{y}) \ge& \log \mathbb{E}_{q(\mbf{W})} \left[\exp(\tilde{\mathcal{L}}_{\mbf{W}}^{\mathrm{lgp}}) \frac{p(\mbf{W})}{q(\mbf{W})} \right] - \mathrm{KL}[q(\mbf{u})||p(\mbf{u})] \\
\ge& \mathbb{E}_{q(\mbf{W})} \log \left[\exp(\tilde{\mathcal{L}}_{\mbf{W}}^{\mathrm{lgp}}) \frac{p(\mbf{W})}{q(\mbf{W})} \right] - \mathrm{KL}[q(\mbf{u})||p(\mbf{u})]
=\mathcal{L}_{\mathrm{slgp}}.
\end{split}
\end{align}
In~\eqref{eq_elbo_lgp_tight}, we could have the independent normal prior $p(\mbf{W}) = \prod_{i=1}^N p(\mbf{w}_i) = \prod_{i=1}^N \mathcal{N}(\mbf{w}_i|\mbf{0}, \mbf{I})$, or more informatively, the independent but expressive prior~\cite{pagnoni2018conditional}
\begin{align} \label{eq_p(w)_mlp}
p(\mbf{w}_i) = \mathcal{N}(\mbf{w}_i|\mathtt{Linear}(\mathtt{MLP}(\mbf{x}_i)), \mathtt{SoftPlus}(\mathtt{MLP}(\mbf{x}_i))),
\end{align}
which amortizes the parameters over a multi-layer perception (MLP) for efficient training; the variational posterior $q(\mbf{W}) = \prod_{i=1}^N q(\mbf{w}_i)$ also takes the Gaussian, with the mean and variance (i) treated as individual hyperparameters~\cite{dutordoir2018gaussian}, or more efficiently, (ii) parameterized as a MLP of both $\mbf{x}$ and $\mbf{y}$, i.e.,
\begin{align} \label{eq_q(w)_mlp}
q(\mbf{w}_i) = \mathcal{N}(\mbf{w}_i|\mathtt{Linear}(\mathtt{MLP}([\mbf{x}_i, y_i])), \mathtt{SoftPlus}(\mathtt{MLP}([\mbf{x}_i, y_i]))).
\end{align}

Moreover, the importance weighted variational inference (IWVI)~\cite{burda2016importance, domke2018importance} further helps pushing the bound towards the marginal likelihood by making the quantity inside the first expectation term of~\eqref{eq_elbo_lgp_tight} concentrated about the mean. This is performed by a sample average over $S$ terms with respect to (w.r.t.) $\mbf{w}$ as
\begin{align}
\mathcal{L}_{\mathrm{slgp}}^{S} = \mathbb{E}_{q(\mbf{W}^{1:S})} \left[\log \frac{1}{S} \sum_{s=1}^S \exp(\tilde{\mathcal{L}}_{\mbf{W}^{s}}^{\mathrm{lgp}})\frac{p(\mbf{W}^{s})}{q(\mbf{W}^{s})}\right] - \mathrm{KL}[q(\mbf{u})||p(\mbf{u})],
\end{align}
where the posterior $q(\mbf{W}^{1:S}) = \prod_{s=1}^S q(\mbf{W}^{s})$, and maximizing the expectation of $p(\mbf{W}^s)/q(\mbf{W}^s)$ w.r.t. $q(\mbf{W}^s)$ represents a regularization: it pushes the posterior $q(\mbf{W}^s)$ towards the prior $p(\mbf{W}^s)$. This bound becomes strictly tighter as $S$ increases, that is, $\mathcal{L}_{\mathrm{slgp}} = \mathcal{L}_{\mathrm{slgp}}^{1} \le 
\cdots \mathcal{L}_{\mathrm{slgp}}^{S} \le \log p(\mbf{y})$~\cite{burda2016importance}, which can be obtained by using Jensen's inequality inversely. From another point of view~\cite{cremer2017reinterpreting}, the IWVI is equivalent to employing a posterior $q_{\mathrm{IW}}(\mbf{W}) = \mathbb{E}_{q(\mbf{W}^{1:S})} \left[\hat{p}(\mbf{y},\mbf{W}) / \left(\frac{1}{S}\frac{\hat{p}(\mbf{y},\mbf{W}^s)}{q(\mbf{W}^s)} \right)\right]$ with $\hat{p}(\mbf{y},\mbf{W}) = \hat{p}(\mbf{y}|\mbf{W}) p(\mbf{W})$. This posterior is more informative than the original $q(\mbf{W})$, and converges to $p(\mbf{W}|\mbf{y})$ when $S \to +\infty$ and $\mbf{u} \to \mbf{f}$.

\textbf{Enabling non-stationarity.} Note that unlike the SHGP and SMGP, the SLGP described so for has no way to tackle non-stationary features. Therefore, as depicted in Fig.~\ref{fig_modulated_gps}(c), we introduce an additional stochastic encoder $p(\mbf{h}|\mbf{w}, \mbf{x}) = p(\mbf{h}|\mbf{w})$\footnote{The latent variable $\mbf{h}$ is actually conditioned on $[\mbf{x}, \mbf{w}]$. But since $\mbf{x}$ is deterministic, we omit it.} as the mapping between the augmented input space $\mathbb{R}^{d_{\mbf{x}}+d_{\mbf{w}}}$ and the latent space $\mathbb{R}^{d_{\mbf{h}}}$ for performing latent representation learning, which thus warps the non-stationary feature to ease the subsequent regression task. The stochastic encoder could be implemented by deep Gaussian process~\cite{salimbeni2019deep}, which however significantly increases model complexity. Alternatively, inspired by the idea of amortized variational inference, we follow~\cite{liu2020deep} and resort to the great representational power of neural networks. Hence, for the stochastic encoder, we take 
\begin{itemize}
	\item the factorized Gaussian prior
	\begin{align}
	p(\mbf{h}|\mbf{w}) = \mathcal{N}(\mbf{h}|\phi(\mbf{x},\mbf{w}), \nu_0 \mbf{I}),
	\end{align}
	where the prior mean $\phi(\mbf{x},\mbf{w})$ represents the mapping of the augmented inputs. Particularly, when $d_{\mbf{h}} = d_{\mbf{x}}+d_{\mbf{w}}$, we have an identity mapping $\phi(\mbf{x},\mbf{w}) = [\mbf{x},\mbf{w}]$; when $d_{\mbf{h}} > d_{\mbf{x}}+d_{\mbf{w}}$, we use the zero-padding strategy $\phi(\mbf{x},\mbf{w}) = [\mbf{x},\mbf{w}, \mbf{0}]$; and finally, when $d_{\mbf{h}} < d_{\mbf{x}}+d_{\mbf{w}}$, we use some dimensionality reduction algorithms, for example, the principle component analysis (PCA), to map $\mbf{w}$ as $\phi(\mbf{x},\mbf{w}) = \mathtt{PCA}([\mbf{x},\mbf{w}])$; and
	\item the variational posterior 
	\begin{align*}
	q(\mbf{h}|\mbf{w}) = \mathcal{N}(\mbf{h}|\bm{\mu}^h, \mathrm{diag}[\bm{\nu}^h])
	\end{align*}
	to be Gaussians factorized over dimensions.The variational mean and variance are parameterized as a MLP of input as
	\begin{align} \label{eq_q(h|w)_mlp}
	\bm{\mu}^h = \mathtt{Linear}(\mathtt{MLP}([\mbf{x},\mbf{w}])), \quad \bm{\nu}^h = \nu_0 \times \mathtt{Sigmoid}(\mathtt{MLP}([\mbf{x},\mbf{w}])),
	\end{align}
	where the shared parameter $\nu_0$ in $p(\mbf{h}|\mbf{w})$ and $q(\mbf{h}|\mbf{w})$ enables knowledge transfer between
	the prior and posterior.
\end{itemize}
Note that though we are using the MLPs for amortized inference, the GP framework alleviates the necessity for fine-tuning or regularization. 

Furthermore, following~\cite{liu2020deep}, a hybrid prior 
\begin{align}
p_{\beta}(\mbf{h}|\mbf{w}) = p^{\beta}(\mbf{h}|\mbf{w}) q^{1-\beta}(\mbf{h}|\mbf{w}), \quad \beta \in [0,1]
\end{align}
is employed in order to arrive at (i) more expressive prior and (ii) adjustable regularization on the latent representation learning $[\mbf{x},\mbf{w}] \to \mbf{h}$. Consequently, we obtain the following ELBO
\begin{align} \label{eq_elbo_lgp_beta_iwvi}
\begin{split}
\mathcal{L}_{\mathrm{slgp}}^{S,\beta} =& \mathbb{E}_{ q(\mbf{H}|\mbf{W}^{1:S})q(\mbf{W}^{1:S})} \left[\log \frac{1}{S} \sum_{s=1}^S \exp(\tilde{\mathcal{L}}_{\mbf{W}^{s}}^{\mathrm{lgp}}) \frac{p_{\beta}(\mbf{H}|\mbf{W}^{s})}{q(\mbf{H}|\mbf{W}^{s})} \frac{p(\mbf{W}^{s})}{q(\mbf{W}^{s})}\right] \\
&- \mathrm{KL}[q(\mbf{u})||p(\mbf{u})] \\
=&\mathbb{E}_{ q(\mbf{H}|\mbf{W}^{1:S})q(\mbf{W}^{1:S})} \left[\log \frac{1}{S} \sum_{s=1}^S \exp(\tilde{\mathcal{L}}_{\mbf{W}^{s}}^{\mathrm{lgp}}) \frac{p^{\beta}(\mbf{H}|\mbf{W}^{s})}{q^{\beta}(\mbf{H}|\mbf{W}^{s})} \frac{p(\mbf{W}^{s})}{q(\mbf{W}^{s})}\right] \\
&- \mathrm{KL}[q(\mbf{u})||p(\mbf{u})],
\end{split}
\end{align}
where $q(\mbf{H}|\mbf{W}^{1:S}) = \prod_{i=1}^N q(\mbf{h}_i|\mbf{w}_i^{1:S}) = \prod_{i=1}^N \prod_{s=1}^S q(\mbf{h}_i|\mbf{w}_i^{s})$. A key advantage of this bound is the capability to tune and control the regularization on latent representation learning through the balance factor $\beta$. As has been investigated in~\cite{liu2020deep}, when $\beta = 1.0$, we pose the complete regularization to constrain the latent representation learning $[\mbf{x},\mbf{w}] \to \mbf{h}$, since maximizing the expectation of $p(\mbf{H}|\mbf{W}^{s})/q(\mbf{H}|\mbf{W}^{s})$ w.r.t. $q(\mbf{H}|\mbf{W}^{s})$ pushes the posterior $q(\mbf{H}|\mbf{W}^{s})$ towards the prior $p(\mbf{H}|\mbf{W}^{s})$; contrarily, the decreasing $\beta$ weakens the regularization to have more powerful representation at the cost of however raising the possibility of over-fitting; particularly, $\beta=0$ offers a \textit{deterministic} encoder for $[\mbf{x},\mbf{w}] \to \mbf{h}$.

\textbf{Tighter bound is not necessarily better.} Though increasing $S$ is found to bring tighter bound, it is however found that the bound $\mathcal{L}_{\mathrm{slgp}}^{S,\beta}$ using $\beta = 1.0$ still often risks severe over-fitting in scenarios with finite number of data points. This may be attributed to the monte carlo approximation quality of $\mathcal{L}_{\mathrm{slgp}}^{S,\beta}$, the maximization of which may not push $q(\mbf{H}|\mbf{W})$ towards $p(\mbf{H}|\mbf{W})$. Empirical evidence has been provided in the numerical experiments in Sec.~\ref{sec_exp}.

It is observed that when $S=1$, the bound~\eqref{eq_elbo_lgp_beta_iwvi} degenerates to the relaxed VI-based ELBO as
\begin{align} \label{eq_elbo_lgp_beta_vi}
\begin{split}
\mathcal{L}_{\mathrm{slgp}}^{\beta} =& \mathbb{E}_{ q(\mbf{H}|\mbf{W})q(\mbf{W})} \left[ \tilde{\mathcal{L}}_{\mbf{W}}^{\mathrm{lgp}} \right] - \beta \mathbb{E}_{q(\mbf{W})} \left[\mathrm{KL}[q(\mbf{H}|\mbf{W})||p(\mbf{H}|\mbf{W})]\right] \\ 
&-\mathrm{KL}[q(\mbf{W})||p(\mbf{W})] - \mathrm{KL}[q(\mbf{u})||p(\mbf{u})],
\end{split}
\end{align}
which now owns an analytical, non-negative KL term to measure the gap between $q(\mbf{H}|\mbf{W})$ and $p(\mbf{H}|\mbf{W})$, the minimization of which pushes the posterior towards the prior. This bound therefore fully utilizes the regularization when $\beta = 1.0$, with however less tight approximation to the objective $\log p(\mbf{y})$, which may harm the performance. 

Therefore, aiming to take advantages from both the IWVI-based ELBO~\eqref{eq_elbo_lgp_beta_iwvi} and the VI-based ELBO~\eqref{eq_elbo_lgp_beta_vi}, we propose a hybrid ELBO as
\begin{align} \label{eq_elbo_lgp_beta_hybrid}
\begin{split}
\mathcal{L}_{\mathrm{slgp}}^{\mathrm{hyb}} =& \mathbb{E}_{ q(\mbf{H}|\mbf{W}^{1:S})q(\mbf{W}^{1:S})} \left[\log \frac{1}{S} \sum_{s=1}^S \exp(\tilde{\mathcal{L}}_{\mbf{W}^{s}}^{\mathrm{lgp}}) \frac{p(\mbf{W}^{s})}{q(\mbf{W}^{s})}\right] \\
&- \beta \mathbb{E}_{q(\mbf{W}^{1:S})} \left[\frac{1}{S} \sum_{s=1}^S \mathrm{KL}[q(\mbf{H}|\mbf{W}^s)||p(\mbf{H}|\mbf{W}^s)]\right] - \mathrm{KL}[q(\mbf{u})||p(\mbf{u})],
\end{split}
\end{align}
which separates the KL term w.r.t. $\mbf{H}$ to achieve the regularized latent representation learning. The hybrid ELBO has higher approximation quality for $\log p(\mbf{y})$ in comparison to~\eqref{eq_elbo_lgp_beta_vi}; meanwhile, it achieves controllable regularization w.r.t. $\mbf{w}$ in comparison to~\eqref{eq_elbo_lgp_beta_iwvi}.

\textbf{Prediction.} Finally, it is notable that when performing prediction at $\mbf{x}_*$, it depends on the posterior $q(\mbf{w}_*)$ which however is unknown. Instead, we take samples from the prior $p(\mbf{w}_*)$ and pass them through the model to output predictive samples. This inconsistency indicates that instead of using the simple unit normal prior $p(\mbf{w})$, we may need to use more informative prior, like~\eqref{eq_p(w)_mlp}, in order to mimic the posterior $q(\mbf{w})$. Consequently, it alleviates the inconsistency when performing prediction. That is why recent works have exploited the usage of expressive priors like mixture of Gaussians and normalizing flow~\cite{bhattacharyya2019conditional}.

\subsection{Discussions}
Table~\ref{tab_comp} summarizes the capability of various GP paradigms together with their time complexity in different scenarios. It is observed that the hemoscedastic, stationary SGP~\cite{hensman2013gaussian} fails in the three challenging scenarios. Contrarily, the three scalable modulated GPs achieve improvement. Among them, the SHGP is not designed for multi-modal scenario, and the Gaussian noise assumption prohibits the modeling of non-Gaussian residuals. Besides, the share of $w(.)$ in~\eqref{eq_hgp} for modulating both the output and noise may weaken the learning of non-stationary features. The time complexity of SHGP is two times that of SGP due to the additional log-GP. As for SMGP, ideally, it is capable of approximating any target distribution, given that we own many GP experts (i.e., a large $T$) and learn the assignment GPs reasonably, which however are quite challenging and raise remarkably high time complexity. As for the final SLGP, it showcases superiority in the following numerical experiments, with the additional complexity $\mathcal{O}(\mathrm{NN})$, which is usually lower than that of SGP, brought by the MLPs for the prior $p(\mbf{w})$ in~\eqref{eq_p(w)_mlp}, and the posteriors $q(\mbf{w})$ in~\eqref{eq_p(w)_mlp} and $q(\mbf{h}|\mbf{w})$ in~\eqref{eq_q(h|w)_mlp}.

\begin{table}
	\caption{The capability and complexity of GP paradigms in various scenarios, with the number of stars indicating the model capability.} 
	\label{tab_comp}
	\centering
	\resizebox{\columnwidth}{!}{%
	\begin{tabular}{lrrrr}
		\hline
		Model & Hetero. noise & Multi. modality & Non-stationarity & Time complexity \\
		\hline
		SGP & $\times$ & $\times$ & $\times$ & $\mathcal{O}(|\mathcal{B}|M^2+M^3)$	\\
		SHGP & $\star$ & $\times$ & $\star$ & $2\mathcal{O}(|\mathcal{B}|M^2+M^3)$	\\
		SMGP & $\star$$\star$ & $\star$$\star$ & $\star$$\star$ & $2T\mathcal{O}(|\mathcal{B}|M^2+M^3)$	\\
		SLGP & $\star$$\star$$\star$ & $\star$$\star$$\star$ & $\star$$\star$$\star$ &$\mathcal{O}(\mathrm{NN})+\mathcal{O}(|\mathcal{B}|M^2+M^3)$	\\		 
		\hline
	\end{tabular}
	}
\end{table}

\section{Related work} \label{sec_related_work}
As for the modeling of heteroscedastic residuals, Goldberg et al~\cite{goldberg1998regression} developed the heteroscedastic GP model $y(\mbf{x}) = f(\mbf{x}) + \mathcal{N}(0, e^{w(\mbf{x})})$, wherein the Gibbs sampling is used to sample from posteriors for model training. The scalability of this model has thereafter been improved through variational inference and distributed learning using sparse approximation~\cite{lazaro2011variational, almosallam2016gpz, liu2020large}. Moreover, in order to tackle non-stationary features and arrive at log-concave likelihood,  the HGP model in~\eqref{eq_hgp} has been first studied in~\cite{ munoz-gonzalez2014divisive, munoz-gonzalez2016laplace} by combining the non-stationary GP $y(\mbf{x}) = e^{w(\mbf{x})} f(\mbf{x}) + \mathcal{N}(0, \nu^{\epsilon})$~\cite{adams2008gaussian} and the above heteroscedastic noise, the inference of which resorts to expectation propagation. The full GP paradigm however makes it only available on small datasets. As an improvement, we here improve the scalability of~\eqref{eq_hgp} under sparse approximation and extend it to the regime of big data; besides, we use variational inference to derive the \textit{closed-form} ELBO~\eqref{eq_elbo_hgp} for efficient model training.

As for the mixture of GPs,\footnote{It belongs to the category of mixture of experts (MoE)~\cite{yuksel2012twenty, masoudnia2014mixture}.} early works seek to infer a gating network $g(.)$, which is usually parameterized as a softmax function, to perform probabilistic separation of the input domain and the assignment of data points, which in turn allow the training of local GP experts~\cite{rasmussen2002infinite, meeds2006alternative, yang2011efficient}. The sparse approximations~\cite{yuan2009variational, sun2011variational, nguyen2014fast} have then been introduced in MGP to reduce the model complexity under the variational expectation maximization framework. The key assumption in the above MGPs is that only one GP expert is used to explain the data at each point. Differently, a whole Bayesian generative model for both the GPs and the assignments has been recently investigated in~\cite{kaiser2019data}, wherein the key assumption is that the data at each position is generated by a number of global GP experts. However, as discussed in Sec.~\ref{sec_mgp}, the discrete multinomial prior $p(\mbf{W}|\mbf{A})$ makes the VI-based ELBO~\eqref{eq_elbo_vi_mgp} hard to use. To sidestep this issue, Kaiser et al.~\cite{kaiser2019data} decided to keep the latent variables $\mbf{W}$ by approximating $\log p(\mbf{y}, \mbf{W})$ rather than the interested $\log p(\mbf{y})$. Contrarily, our work devotes to marginalize all the latent variables out to \textit{directly} approximate the marginal likelihood $\log p(\mbf{y})$ and derive a \textit{tighter} bound for model training with higher quality.

As for the latent GP, it was first proposed in~\cite{wang2012gaussian} for modeling hereroscedastic non-Gaussian residuals, and then extended for conditional density estimation~\cite{bodin2017latent, dutordoir2018gaussian, salimbeni2019deep} and disentangled learning~\cite{martens2019decomposing}. The similar idea of augmenting inputs with latent variables was also independently presented from the regime of reinforcement learning~\cite{depeweg2019learning}. It is also notable that the LGP is close to the Gaussian process latent variable model (GPLVM)~\cite{lawrence2005probabilistic,damianou2014variational}, despite that the latter seeks to infer the low-dimensional latent variables for unsupervised learning. Besides, the LGP has the same model structure to the conditional variational autoencoder (CVAE)~\cite{sohn2015learning}, though they were developed from different views. Our SLGP model is most related to the work in~\cite{dutordoir2018gaussian}, which combines the amortized variational inference with LGP for conditional density estimation. The major differences are that (i) our SLGP introduces an additional, \textit{regularized stochastic encoder} $q(\mbf{h}|\mbf{w})$ to enhance the power of latent representation; and (ii) we derive a \textit{hybrid} and \textit{tighter} ELBO to achieve high training quality.

\section{Numerical experiments} \label{sec_exp}
This section comprehensively investigates the characteristics and performance of different scalable modulated GPs for approximating diverse distributions on three toy cases, eight UCI datasets, the large New York City Taxi dataset, and finally the \texttt{mnist} image generation task. All the experiments are performed on a Linux workstation with eight 3.20 GHz cores, nvidia GTX1080Ti, and 32GB memory.

\subsection{Toy cases}
This section employs three toy cases with different characteristics to showcase the performance of modulated GPs. The first case 
\begin{align*}
y(x) = \cos(5x) \exp(-0.5x) + 0.25\cos(6x+1) \exp(-x) \mathcal{N}(0,1), \quad x \in [-2, 2]
\end{align*}
has the heteroscedastic Gaussian noise; the second case is a non-stationary step function around $x=0.5$ with a constant noise $\epsilon \sim \mathcal{N}(0, 1e^{-4})$; and finally the third case is the multi-modal \texttt{moon} data imported from the \texttt{scikit-learn} package~\cite{scikit-learn}. We generate $n = 1000$ training points for the heteroscedastic case, $n = 500$ for the step case, and $n=200$ for the multi-modal case, and predict at 500 evenly distributed points. 

\begin{figure}[t!]
	\centering
	\includegraphics[width=1.0\textwidth]{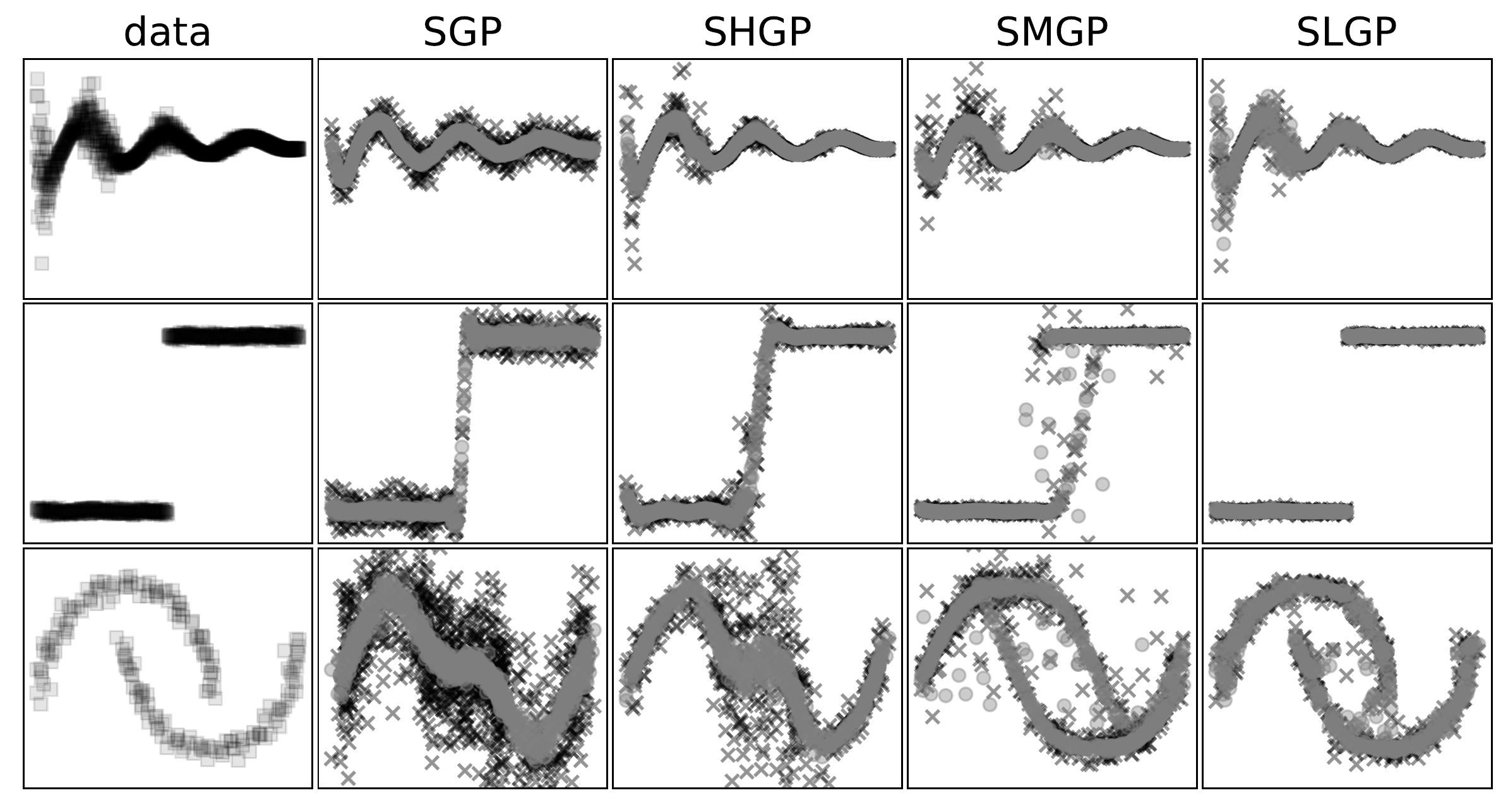}
	\caption{Single samples from the predictions of various GPs on three toy cases. The first row represents the training points for the three cases. As for the remaining rows, the black crosses represent the samples from $p(y_*|\mbf{y})$, while the gray circles are the samples from $p(f_*|\mbf{y})$.}
	\label{fig_toy}
\end{figure}

In the comparison study, the training data has been normalized before model training. The modulated GPs use the squared exponential (SE) kernel, and $m = 50$ inducing variables. Particularly, the SMGP uses $K=4$ GP experts; the SLGP uses the balance parameter $\beta = 0.01$ and a single latent input (i.e., $d_{\mbf{w}}=1$), and adopts the MLPs for $p(\mbf{w})$, $q(\mbf{w})$ and $q(\mbf{h}|\mbf{w})$ with three hidden layers, each of which owns 100 units and takes the ReLU activation function. Note that both SMGP and SLGP use $S=10$ MC samples to evaluate their ELBOs for model training. As for the optimization, we employ the Adam solver with the learning rate of $0.005$ and run it over 10000 iterations. Fig.~\ref{fig_toy} illustrates the prediction samples generated by the three modulated GPs. The hemoscedastic, stationary SGP~\cite{hensman2013gaussian} is also included as a baseline. We conclude the following findings from the comparative results.

\begin{figure}[t!]
	\centering
	\includegraphics[width=1.0\textwidth]{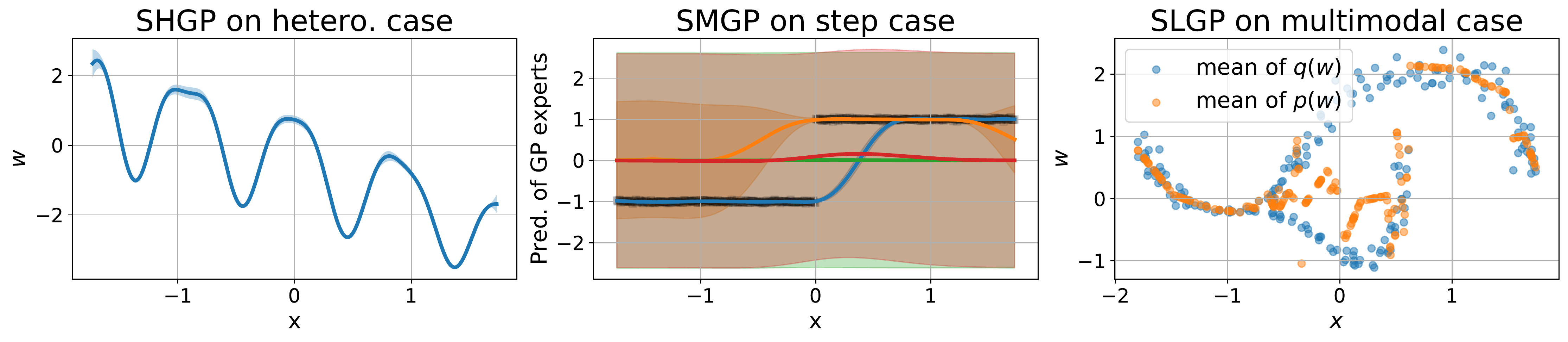}
	\caption{Illustration of, from left to right, (i) the modulation function $w(.)$ learnt by SHGP on the heteroscedastic case, (ii) the predictions of four GP experts learnt by SMGP on the step case, and finally (iii) the mean of $q(\mbf{w})$ and $p(\mbf{w})$ learnt by SLGP at training points on the multi-modal case. Note that the shaded region represents 95\% confidence interval of the prediction, and the inputs of the three cases have been normalized in the plots.}
	\label{fig_W_toy}
\end{figure}

\textbf{The modulated GPs outperform the SGP.} The homoscedastic SGP is only able to estimate the simple constant noise in data, and has no way to tackle non-stationary and multi-modal features. As a result, the generated samples cannot well recover the three data distributions. Contrarily, the modulated GPs approximate the data more accurately by extracting richer statistical behavior behind data via the modulation $\mbf{w}$. As shown in Fig.~\ref{fig_W_toy}, (i) the modulation variables learnt by SHGP capture the varying noise in the heteroscedastic case; (ii) the four GP experts learnt by SMGP describe part of the data for capturing the step behavior; and (iii) the mean of $q(\mbf{w})$ learnt by SLGP encodes the multi-modality of the \texttt{moon} case, and the expressive prior $p(\mbf{w})$ mimics the posterior for better prediction.

\textbf{The SLGP is promising in various scenarios.} The SHGP uses an additional latent function $w(.)$ to describe the varying Gaussian noise and calibrate the amplitude of output. Hence, it perfectly approximates the first case and improves over SGP in capturing the step case. Due to the Gaussian noise assumption, the SHGP however cannot fit the multi-modal case well. The SMGP describes the distributions more flexibly by mixing several GP experts. The proper number $T$ of experts however is problem-dependent. Ideally, the SMGP improves with $T$; however, it meanwhile toughens the model training due to the exponentially increased number of hyperparameters and the more complicated model structure. In comparison to the flexible but average performance of SMGP, the SLGP is promising. By including stochasticity in the augmented input space and using an NN-assisted stochastic encoder, the SLGP well recovers the three distributions. 

\textbf{The SLGP may sacrifice the homoscedastic noise.} It is found that the GP framework uses the estimated variance $\nu^f_*$ to quantify the \textit{epistemic} uncertainty in model parameters~\cite{kendall2017what}. As for the \textit{aleatoric} uncertainty inherited in the observations, the SHGP employs an additional GP to fit it; while the SGP, SMGP and SLGP use a simple \textit{i.i.d} noise $\mathcal{N}(0, \nu^{\epsilon})$ for approximation. It is observed from Fig.~\ref{fig_toy} that the latent samples of SLGP from $p(f_*|\mbf{y})$ are almost the same as the samples from $p(y_*|\mbf{y})$ for the first heteroscedastic case, which indicates that the SLGP here sacrifices the \textit{i.i.d.} noise. That is, the SLGP pushes the noise variance towards zero, e.g., it has the pretty small estimation $\nu^{\epsilon} = 0.0043$. This is because for SLGP, the \textit{i.i.d.} noise is inconsistent to the heteroscedastic property of the first case. Thus, it tends to ignore the noise term and resorts to the $p(f_*|\mbf{y})$ conditioned on augmented stochastic inputs to explain the data.

\subsection{UCI regression}
This section further investigates the performance of the three modulated GPs on eight UCI benchmarks~\cite{Dua:2019} summarized in Table~\ref{tab_uci_data}. In addition to the SGP~\cite{hensman2013gaussian}, the comparative study adopts the GP competitors including 
\begin{itemize}
\item the stochastic variational HGP (SVHGP)~\cite{liu2020large} which uses the latent function $w(.)$ to only modulate the noise variance,
\item the data association GP (DAGP)~\cite{kaiser2019data} which maximizes the joint likelihood $p(\mbf{y}, \mbf{W})$ rather than the marginal likelihood $p(\mbf{y})$ for model training, and
\item the GP conditional density estimation (GPCDE)~\cite{dutordoir2018gaussian} which derives an optimal ELBO for LGP.
\end{itemize}
Besides, the competitors also include the neural network counterparts including 
\begin{itemize}
\item the heteroscedastic neural network (HNN)~\cite{nix1994estimating} with the last layer having two outputs, one for the prediction mean of a Gaussian and the other for the uncertainty of the prediction mean,
\item the mixture of density network (MDN)~\cite{bishop1994mixture} with the last layer outputting the means and variances of multiple Gaussians together with the related coefficients to mix them, and finally,
\item the conditional variational auto-encoder(CVAE)~\cite{sohn2015learning} which has the similar structure to the proposed SLGP with the main difference being that it is a pure NN architecture without stochastic regularization.
\end{itemize}
The model configurations on the UCI datasets are elaborated in Appendix~\ref{app_exp_detail}. Note that the SLGP performs grid search of the balance parameter $\beta$ from the candidate set $\{1.0, 0.5, 0.1, 0.01\}$. Table~\ref{tab_uci} reports the average comparative results over ten runs in terms of negative log likelihood (NLL), which is estimated by the kernel density estimator built from the prediction samples. We have the following findings from the comparative results.

\begin{table}
	\caption{Eight UCI benchmark datasets.} 
	\label{tab_uci_data}
	\centering
	%\resizebox{\columnwidth}{!}{%
	\begin{tabular}{lrrr}
		\hline
		dataset & $N_{\mathrm{train}}$ & $N_{\mathrm{test}}$ & $d_{\mbf{x}}$ \\
		\hline
		\texttt{boston} & 456 & 50 &13	\\
		\texttt{energy} & 692 & 76 & 8	\\
		\texttt{concrete} & 927 & 103 & 8	\\
		\texttt{wine-red} & 1440 & 159 & 11\\ 	
		\texttt{kin8nm} & 7373 & 819 & 8 \\	
		\texttt{power} & 8612 & 956 & 4	\\
		\texttt{naval} & 10741 & 1193 & 16	\\	
		\texttt{protein} & 41157 & 4573 & 9	\\			 
		\hline
	\end{tabular}
	%}
\end{table}

\begin{sidewaystable}
	\caption{The negative log likelihood (NLL) results of modulated GPs and their GP (marked in orange) and NN (marked in blue) counterparts on the UCI datasets, with the best and second-best results marked in gray and light gray, respectively. For the NLL criterion, lower is better. Besides, the last row provides the optimal balance parameter $\beta$ of SLGP through grid search.} 
	\label{tab_uci}
	\centering
	%\resizebox{\columnwidth}{!}{%
		\begin{tabular}{lrrrrrrrr}
			\hline
			model &\texttt{boston} &\texttt{energy} &\texttt{concrete} &\texttt{wine-red} &\texttt{kin8nm} &\texttt{power} &\texttt{naval} &\texttt{protein} \\
			\hline
			\hline
			SGP &2.3325$_{\pm0.0723}$ &1.4431$_{\pm0.0331}$ &3.0514$_{\pm0.0313}$ &0.9527$_{\pm0.0544}$ &-0.9947$_{\pm0.0135}$ &2.7274$_{\pm0.0097}$ &-7.0355$_{\pm0.6860}$ &2.8784$_{\pm0.0080}$ \\
			\hline
			{\color{blue}HNN} & 3.5959$_{\pm0.2348}$ &2.0527$_{\pm0.3290}$ &2.7110$_{\pm0.1831}$ &1.0314$_{\pm0.1625}$ &1.4230$_{\pm0.5217}$ &2.5078$_{\pm0.0411}$ &-5.2414$_{\pm0.7426}$ &2.1944$_{\pm0.1062}$ \\
			{\color{orange}SVHGP} &\cellcolor{mygray}2.1863$_{\pm0.0530}$ &\cellcolor{mygray}1.0108$_{\pm0.0786}$ &2.9546$_{\pm0.0266}$ &0.8585$_{\pm0.0652}$ &\cellcolor{mygray}-1.0413$_{\pm0.0118}$ &2.6678$_{\pm0.0136}$ &-7.1000$_{\pm0.3524}$ &2.7445$_{\pm0.0130}$ \\
			SHGP &\cellcolor{lightgray}2.1174$_{\pm0.0678}$	&1.0762$_{\pm0.0774}$ &2.9706$_{\pm0.0307}$ &0.8991$_{\pm0.0579}$ &-1.0378$_{\pm0.0120}$ &2.7019$_{\pm0.0137}$ &\cellcolor{lightgray}-7.4956$_{\pm0.1271}$ &2.7628$_{\pm0.0114}$ \\
			\hline
			{\color{blue}MDN} &3.3399$_{\pm0.1768}$ &1.3182$_{\pm0.2403}$ &2.6500$_{\pm0.1326}$ &0.3965$_{\pm0.9257}$ &1.7145$_{\pm0.2523}$ &2.4702$_{\pm0.0252}$ &-3.5556$_{\pm1.0886}$ &\cellcolor{mygray}1.9195$_{\pm0.0219}$ \\
			{\color{orange}DAGP} &2.3004$_{\pm0.0535}$ &1.1484$_{\pm0.0938}$ &3.0369$_{\pm0.0377}$ &0.9533$_{\pm0.0621}$ &-0.9985$_{\pm0.0176}$ &2.7186$_{\pm0.0087}$ &-7.2886$_{\pm0.4019}$ &2.8367$_{\pm0.0094}$ \\
			SMGP &2.2094$_{\pm0.0770}$ &1.1500$_{\pm0.1117}$ &2.9681$_{\pm0.0321}$ &1.1342$_{\pm0.0846}$ &-1.0263$_{\pm0.0128}$ &2.7001$_{\pm0.0110}$ &\cellcolor{mygray}-7.3514$_{\pm0.1601}$ &2.7529$_{\pm0.0118}$ \\
			\hline
			{\color{blue}CVAE} &3.6231$_{\pm0.2309}$ &1.7857$_{\pm0.1825}$ &\cellcolor{mygray}2.6141$_{\pm0.1813}$ &\cellcolor{lightgray}-0.7183$_{\pm0.3034}$ &1.3869$_{\pm0.2132}$ &\cellcolor{lightgray}2.3990$_{\pm0.0472}$ &-5.8454$_{\pm0.3416}$ &1.9666$_{\pm0.0237}$ \\
			{\color{orange}GPCDE} &2.3292$_{\pm0.0662}$ &1.2102$_{\pm0.0636}$ &3.0572$_{\pm0.0339}$ &0.9516$_{\pm0.0602}$ &-0.9927$_{\pm0.0136}$ &2.7293$_{\pm0.0083}$ &-6.9172$_{\pm0.2030}$ &2.8181$_{\pm0.0053}$ \\
			SLGP &2.2075$_{\pm0.0804}$ &\cellcolor{lightgray}0.6821$_{\pm0.3064}$ &\cellcolor{lightgray}2.5031$_{\pm0.1559}$ &\cellcolor{mygray}0.6663$_{\pm0.0904}$ &\cellcolor{lightgray}-1.0414$_{\pm0.0135}$ &\cellcolor{lightgray}2.4083$_{\pm0.0368}$ &-6.3133$_{\pm0.2206}$ &\cellcolor{lightgray}1.9089$_{\pm0.0311}$ \\
			$\beta$ &1.0 &1.0 &0.5 & 1.0 &0.5 &0.1 &0.1 &0.01 \\
			\hline
		\end{tabular}
	%}
\end{sidewaystable}

\textbf{The SLGP generally outperforms the others.} Similar to the results on toy cases, the SLGP again showcases superiority on six out of the eight UCI datasets. Besides, it is observed in Table~\ref{tab_uci} that the optimal balance parameter $\beta$ is problem-dependent. As for the first four small datasets, the SLGP prefers using $\beta=1$ in order to fully utilize the regularization to guard against over-fitting. Contrarily, the SLGP decides to use a small $\beta$ on the remaining four large datasets for improving the latent representation learning, which is beneficial for extracting the underlying patterns under massive data. As for the other two modulated GPs, they outperform the SGP for most of the cases. Finally, the mixture of GPs, which ideally has high flexibility and capability, does not show significant superiority over the SHGP. This may be attributed to the difficulty in training such complex model.

\textbf{The modulated GPs usually outperform the GP counterparts.} As for SLGP, it performs better than GPCDE on all the eight benchmarks, due to the regularized latent representation learning $q(\mbf{h}|\mbf{w})$ together with the hybrid and tight ELBO~\eqref{eq_elbo_lgp_beta_hybrid}. As for SMGP, it outperforms the DAGP in six out of the eight benchmarks due to the more reasonable ELBO~\eqref{eq_elbo_mgp}, which marginalizes all the latent variables out. As for SHGP, however, it does not show superiority over SVHGP by additionally modulating the amplitude of output $f(.)$ in~\eqref{eq_hgp}. As has been explained before, this may be attributed to the share of $w(.)$ in SHGP for modulating both the output and noise, which may weaken the learning of non-stationary features. 

\textbf{The NN counterparts suffer from over-fitting on small datasets.} It is not surprising to observe that in comparison to the modulated GPs, the NN counterparts without regularization risk over-fitting on small datasets, for example, the \texttt{boston} and \texttt{energy} datasets. This issue can be alleviated by some fine-tuning tricks, e.g., model pruning, adaptive learning rate, early stop, dropout, and data augmentation. The NN models, especially the CVAE, perform well on the remaining large datasets, because the many data points themselves act as a regularization. Contrarily, though being a hybrid model of NN and GP, the well-calibrated SLGP under the GP framework yields robust and superior performance on both small and large datasets without fine-tuning tricks.

\begin{figure}[t!]
	\centering
	\includegraphics[width=1.0\textwidth]{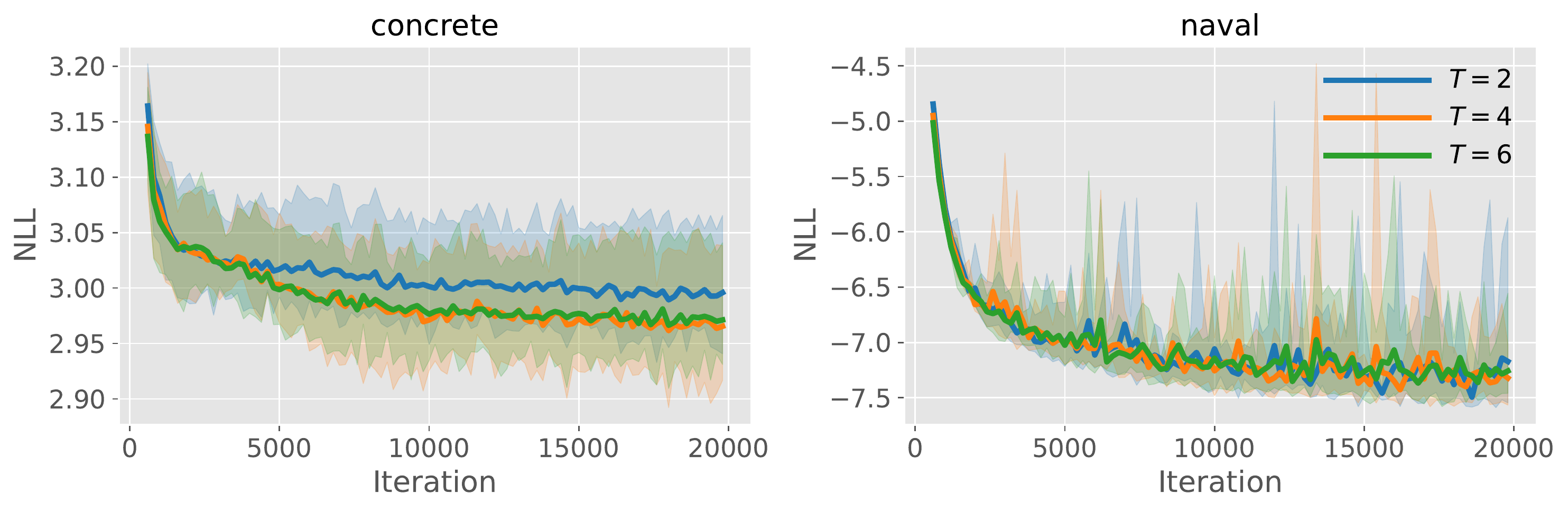}
	\caption{Impact of parameter $T$ on the performance of SMGP. The curve represents the mean over ten runs, while the shaded region represents the min/max bound at each iteration.}
	\label{fig_smgp_k}
\end{figure}

\textbf{The SMGP improves with $T$?} The mixture of more GP experts in SMGP is expected to improve the capability of fitting more complicated data distributions. The comparative results in Fig.~\ref{fig_smgp_k} on the small \texttt{concrete} and large \texttt{naval} datasets however cannot strongly support this claim. It is observed that the SMGP using $T=4$ improves over that with $T=2$ on the \texttt{concrete} dataset; but the larger $T=6$ brings almost no improvement. More seriously, the SLMP is insensitive to $T$ on the \texttt{naval} dataset. This again confirms that training the SMGP comprising $2 \times T$ GP experts with interactions is a challenging task.

\begin{figure}[t!]
	\centering
	\includegraphics[width=1.0\textwidth]{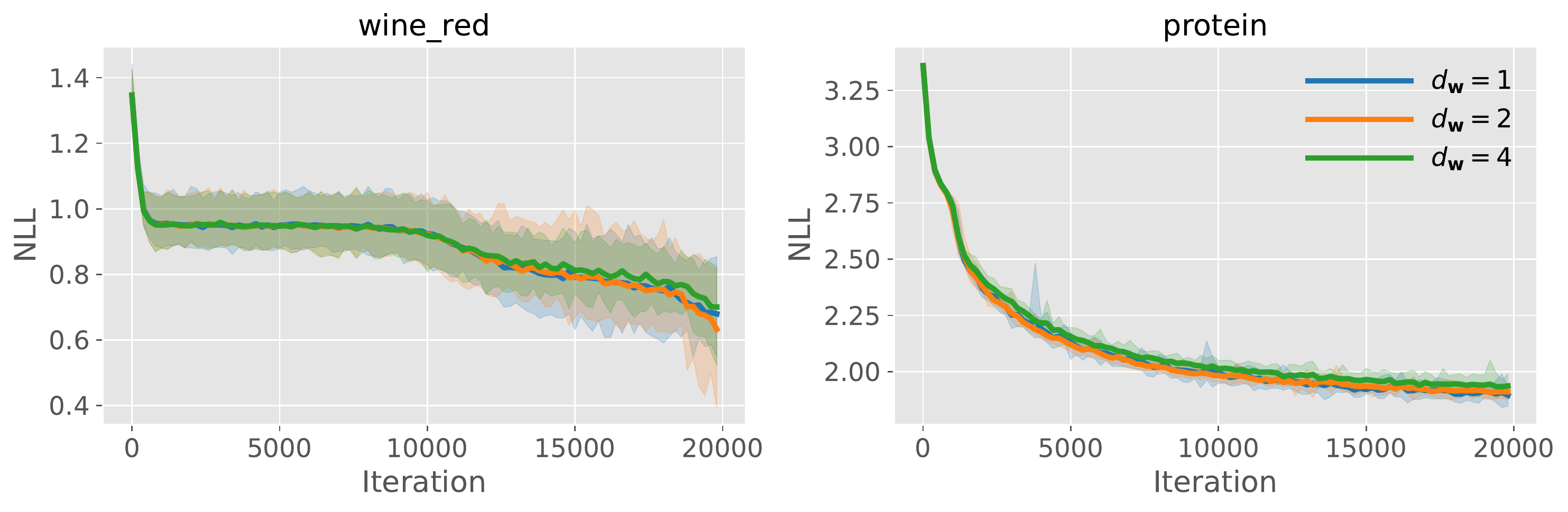}
	\caption{Impact of $d_{\mbf{w}}$ on the performance of SLGP. The curve represents the mean over ten runs, while the shaded region represents the min/max bound at each iteration.}
	\label{fig_slgp_w}
\end{figure}

\textbf{The SLGP improves with $d_{\mbf{w}}$?} Fig.~\ref{fig_slgp_w} investigates the impact of the dimensionality $d_{\mbf{w}}$ of the latent inputs $\mbf{w}$ on the performance of SLGP on the small \texttt{wine-red} and large \texttt{protein} datasets. It is found that the performance of SLGP is insensitive to $d_{\mbf{w}}$ on the two datasets. This may be attributed to the sufficient capability of $d_{\mbf{w}} = 1$ for describing statistical structures on the two dataset. We again investigate the impact of $d_{\mbf{w}}$ on the challenging \texttt{mnist} image dataset in Sec.~\ref{sec_mnist} and observe the benefits brought by large $d_{\mbf{w}}$.

\begin{figure}[t!]
	\centering
	\includegraphics[width=1.0\textwidth]{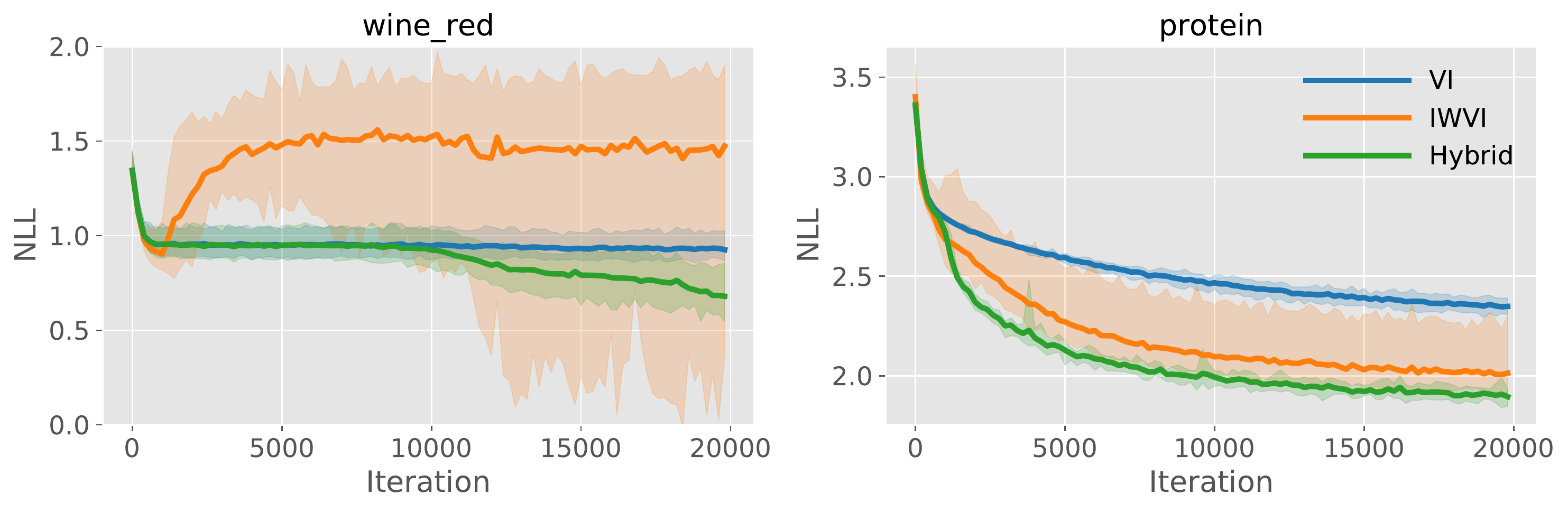}
	\caption{Impact of inference strategy on the performance of SLGP. The curve represents the mean over ten runs, while the shaded region represents the min/max bound at each iteration.}
	\label{fig_slgp_vi}
\end{figure}

\textbf{The hybrid inference in SLGP outperforms VI and IWVI.} As stated before, the IWVI helps derive a tighter bound~\eqref{eq_elbo_lgp_beta_iwvi} than the commonly used VI-based bound~\eqref{eq_elbo_lgp_beta_vi}. The bound is indeed tighter at the cost of however losing the explicit control of regularization on the stochastic encoder when resorting to the MC approximation. This risks severe over-fitting on the small \texttt{wine-red} dataset even when we are using $\beta = 1.0$, as shown in Fig.~\ref{fig_slgp_vi}. Differently, the hybrid bound~\eqref{eq_elbo_lgp_beta_hybrid} takes the advantages of both the VI- and IWVI-based ELBOs. Consequently, it outperforms the VI-based inference on the small \texttt{wine-red} dataset, and even performs better than the IWVI-based inference on the large \texttt{protein} dataset.

\subsection{Large-scale spatio-temporal modeling}
This section applies the three scalable modulated GPs to the large-scale 2016 New York City Taxi (\texttt{nytaxi}) dataset, which is composed of the records of more than 1.4 million taxi trips.\footnote{The data is available at \url{https://www.kaggle.com/c/nyc-taxi-trip-duration/data}.} We employ the \texttt{Bayesian Benchmarks}\footnote{\url{https://github.com/hughsalimbeni/bayesian_benchmarks}.} package to choose the trips within the Manhattan area. We attempt to predict the duration of trip with respect to the drop-off location, the pick-up location, the day of week, and the time of day. It is notable that the day of week and the time of day are transformed as sine and cosine with the natural periods, thus resulting in totally eight inputs. The model configurations on this experiment are elaborated in Appendix~\ref{app_exp_detail}.

\begin{figure}[t!]
	\centering
	\includegraphics[width=.6\textwidth]{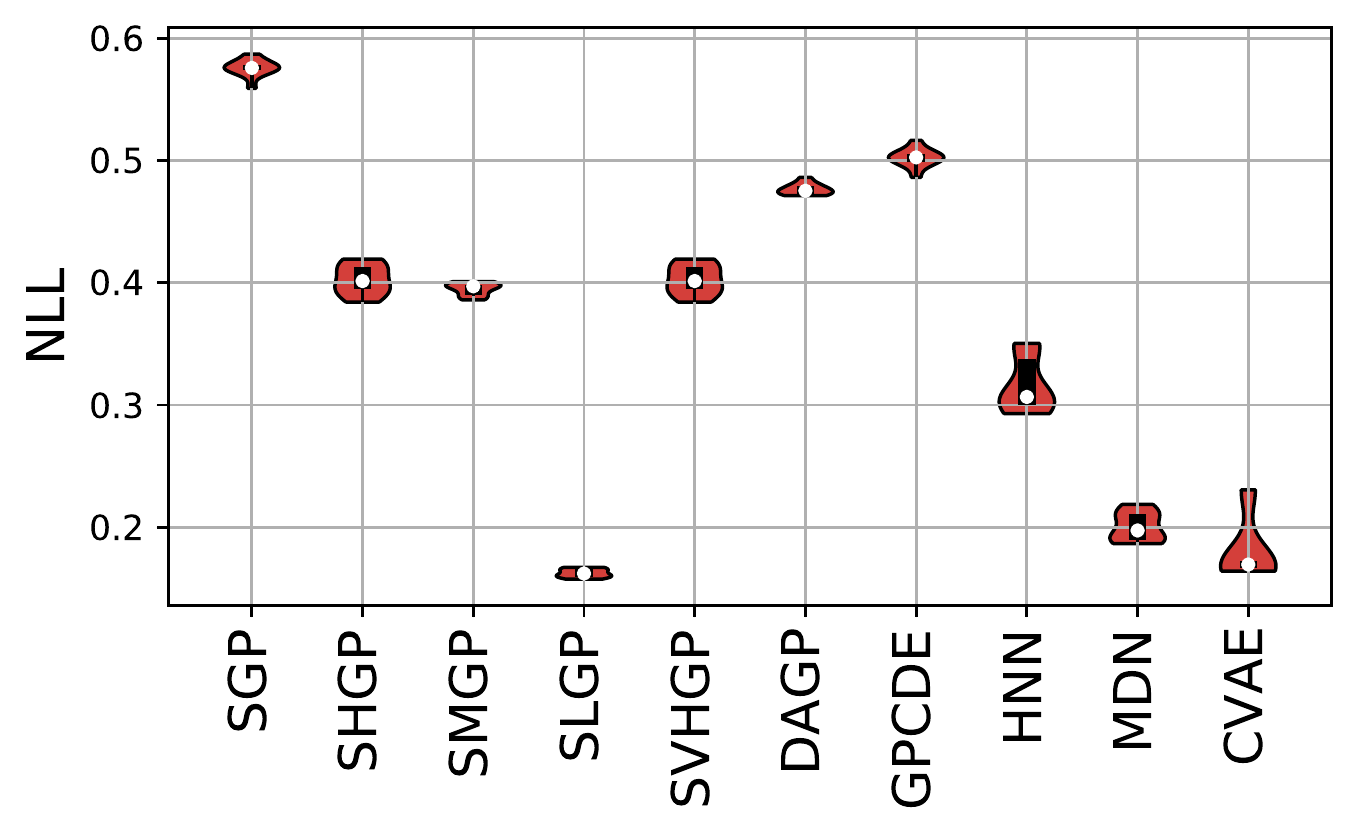}
	\caption{The negative log likelihood (NLL) results of modulated GPs and their GP and NN counterparts on the large-scale \texttt{nytaxi} dataset.}
	\label{fig_nytaxi_violinplot}
\end{figure}

\begin{figure}[t!]
	\centering
	\includegraphics[width=1.0\textwidth]{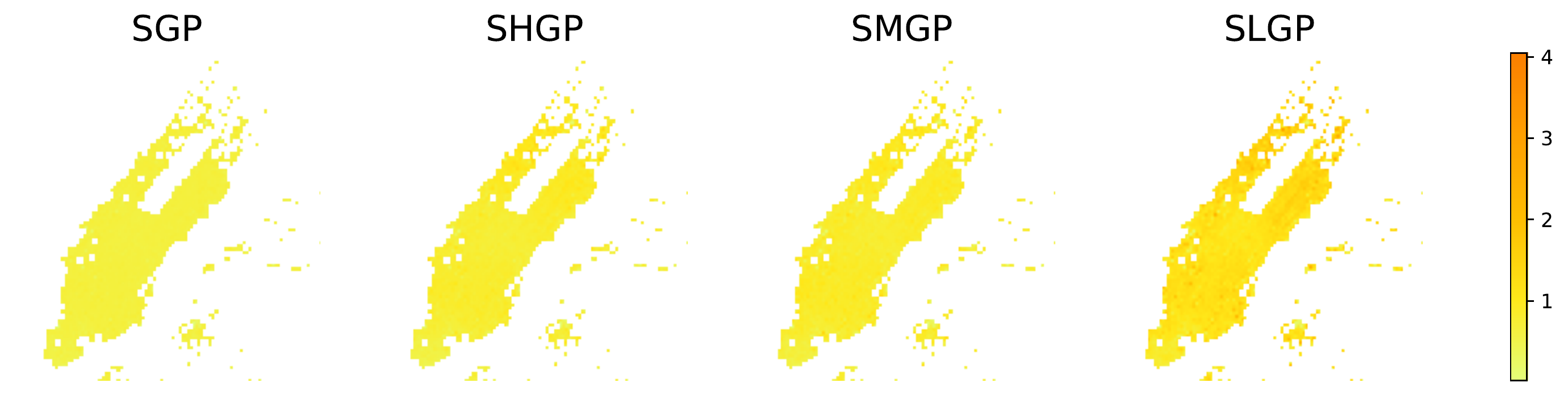}
	\caption{Illustration of the prediction densities of SGP, SHGP, SMGP and SLGP estimated by the \texttt{KernelDensity} method from the \texttt{scikit-learn} package~\cite{scikit-learn} w.r.t. the pick-up locations.}
	\label{fig_nytaxi_density}
\end{figure}

Fig.~\ref{fig_nytaxi_violinplot} shows the comparative results of modulated GPs and their counterparts over ten runs in terms of NLL. It is observed that all the three modulated GPs significantly outperform the conventional SGP. Specifically, the SMGP performs slightly better than the SHGP; the SMGP and SLGP outperform their GP counterparts (DAGP and GPCDE), but the SHGP and SMGP perform worse than their NN counterparts on this large data; finally, the SLGP shows remarkable superiority on this large dataset, even in comparison to CVAE. Furthermore, Fig.~\ref{fig_nytaxi_density} illustrates the estimated densities of the four GPs with respect to the pick-up locations. It is observed that the three modulated GPs, especially the SLGP, have higher likelihood to explain the data.

\subsection{Image generation task} \label{sec_mnist}
Finally, since the powerful SLGP and its NN counterpart CVAE~\cite{sohn2015learning} showcase superiority on the foregoing numerical experiments, this section compares them on the high-dimensional \texttt{mnist} dataset that is composed of 70000 hand-written digit images for density estimation as well as sample generation.\footnote{The dataset is available at \url{http://yann.lecun.com/exdb/mnist/}.} We choose 90\% of the dataset as training data, and the remaining 10\% for testing. Different from the above tasks, the image task has ten-dimensional inputs encoded from the one-hot labels, and 784 vectorized pixel outputs. Appendix~\ref{app_exp_detail} describes the model settings as well as the MLP architectures of SLGP and CVAE. In this experiment, we additionally investigate the impact of latent dimensionality by using $d_{\mbf{w}} = 2$ and $d_{\mbf{w}} = 16$, respectively.

\begin{table}
	\caption{The negative log likelihood (NLL) results on the \texttt{mnist} dataset. For this criterion, lower is better.} 
	\label{tab_mnist}
	\centering
	%\resizebox{\columnwidth}{!}{%
	\begin{tabular}{lrr}
		\hline
		$d_{\mbf{w}}$ & CVAE &SLGP \\
		\hline
		2 & -825.1102	&\textbf{-1198.2542} \\
		16 & -935.4681	&\textbf{-1225.2195} \\	 
		\hline
	\end{tabular}
	%}
\end{table}

\begin{figure*}[t!]
	\centering
	\begin{subfigure}
		\centering
		\includegraphics[width=0.49\textwidth]{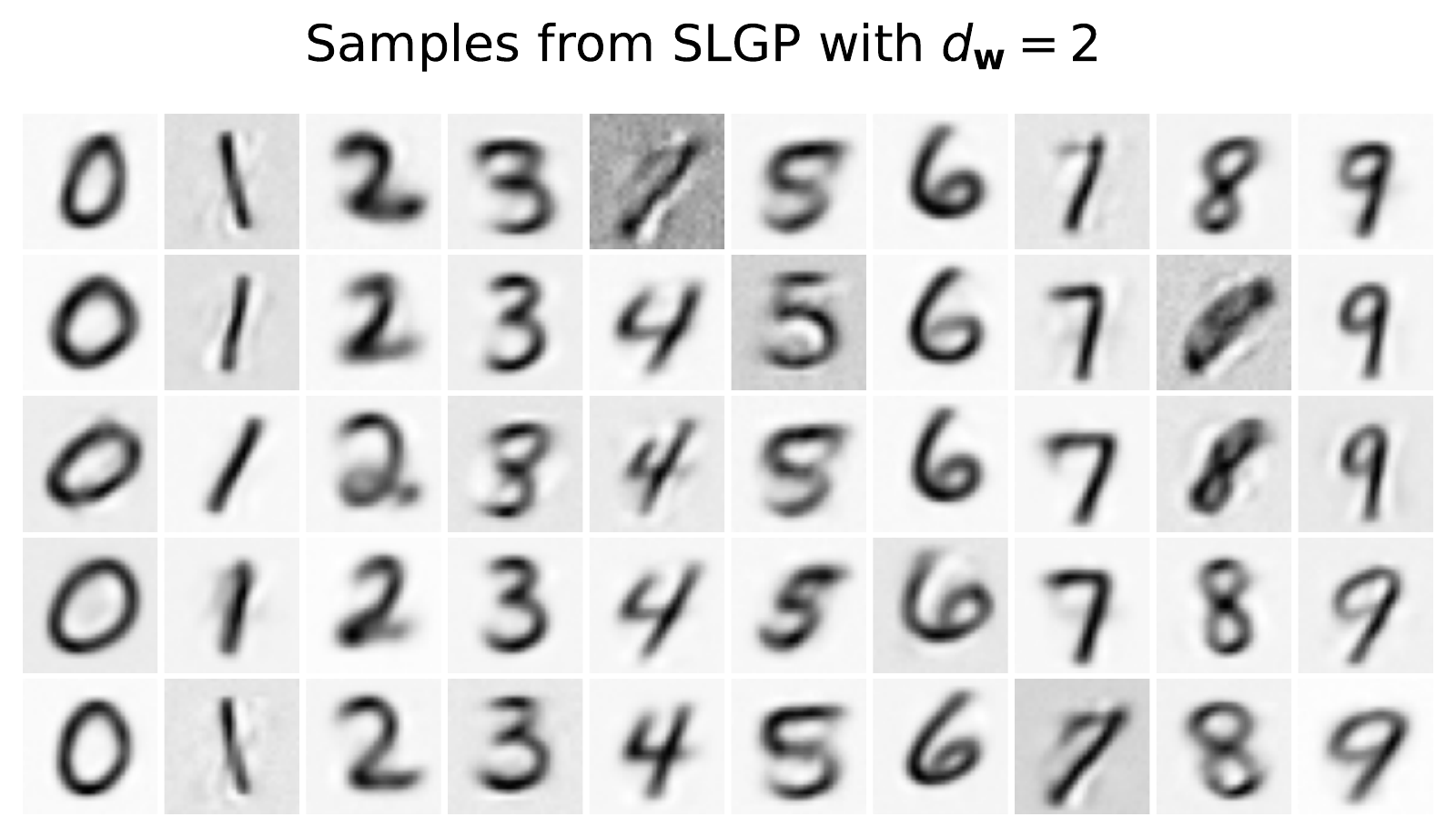}
	\end{subfigure}%
	\begin{subfigure}
		\centering
		\includegraphics[width=0.49\textwidth]{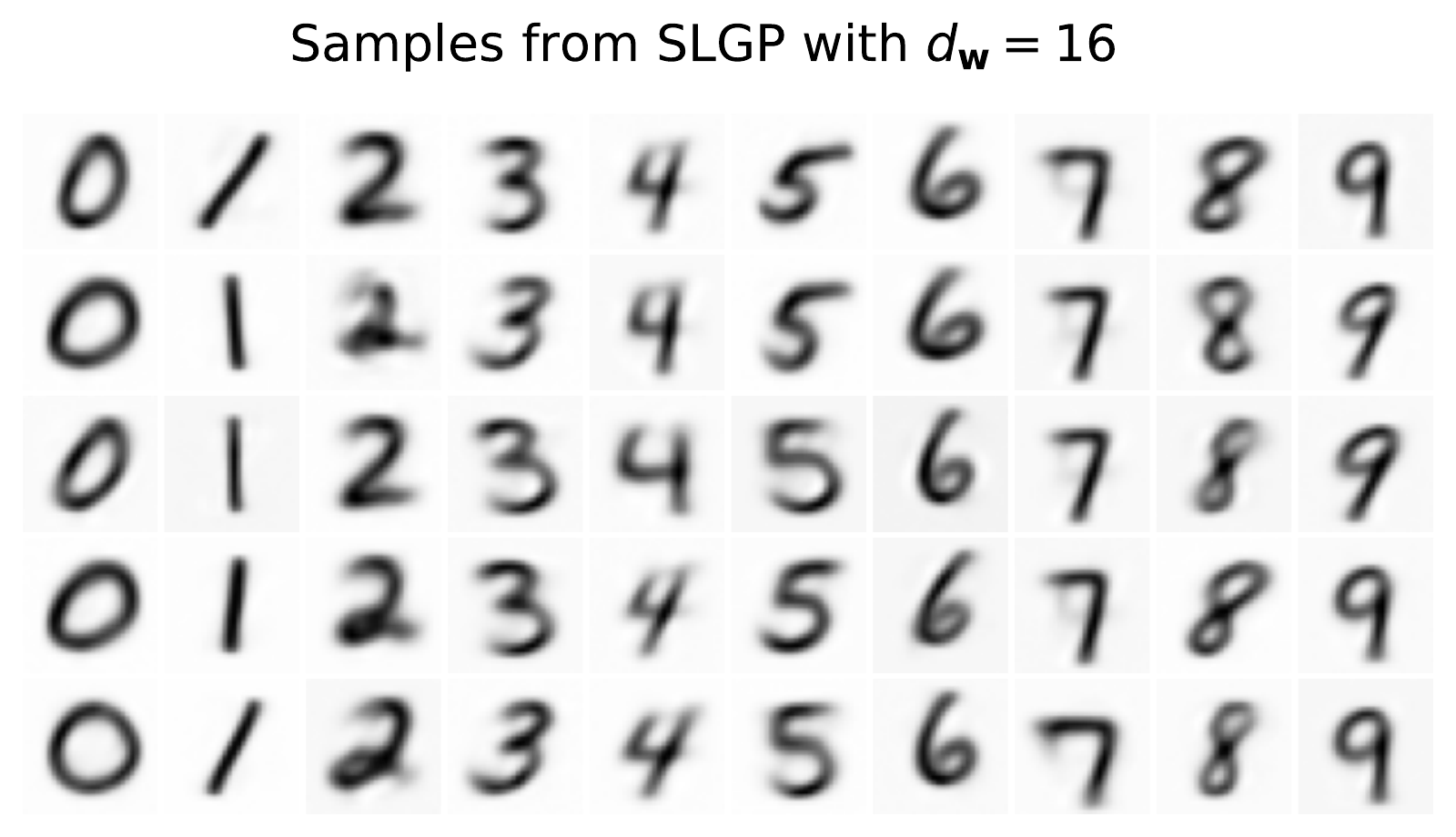}
	\end{subfigure} 
	\caption{The 0-10 digit images sampled from the distributions learnt by SLGP with $d_{\mbf{w}} = 2$ and $d_{\mbf{w}} = 16$, respectively.}
	\label{fig_slgp_mnist} 
\end{figure*}

Table~\ref{tab_mnist} reports the NLL results of CVAE and SLGP on the \texttt{mnist} dataset. It is observed that the Bayesian framework helps the SLGP outperform the CVAE, and both of them improve with the latent dimensionality $d_{\mbf{w}}$. Besides, Fig.~\ref{fig_slgp_mnist} depicts the 0-10 digit images sampled from the distributions learnt by SLGP using $d_{\mbf{w}} = 2$ and $d_{\mbf{w}} = 16$, respectively. It is observed that the samples generated by the SLGP with $d_{\mbf{w}} = 2$ have obviously noisy background and misidentify some digits, for example, '1' and '7'. By increasing $d_{\mbf{w}}$ up to sixteen, we are now capable of generating clear and diverse digit images conditioned on the labels.

\section{Conclusion} \label{sec_conclusion}
In order to enhance the capability of learning rich statistical representation for GP from massive data, we studied three scalable modulated GPs, which encode challenging details into the latent variable $\mbf{w}$ by modulating the outputs or inputs. The variational inference helps further derive analytical or tighter ELBOs for efficient and effective model training. We extensively evaluate these scalable modulated GPs and compare them against state-of-the-art GP and NN counterparts on various tasks. It is observed that (i) they outperform the conventional scalable GP in terms of the quality of predictive distribution in various challenging scenarios, and (ii) particularly, the SLGP shows remarkable performance in learning various distributions at the cost of however may sacrificing the homoscedastic noise in some circumstances.

\section*{Acknowledgments}
This work was supported by the Fundamental Research Funds for the Central Universities (DUT19RC(3)070) at Dalian University of Technology, and it was partially supported by the Research and Innovation in Science and Technology Major Project of Liaoning Province (2019JH1-10100024), the National Key Research and Development Project (2016YFB0600104), and the MIIT Marine Welfare Project (Z135060009002).

\begin{appendices}
\section{Acronyms and notations} \label{app_notation}
\subsection{Acronyms}
$
\begin{array}{ll}
    \mbox{CVAE} & \mbox{Conditional Variational Autoencoder} \\
    \mbox{DAGP} & \mbox{Data Association Gaussian Process} \\
	\mbox{ELBO} & \mbox{Evidence Lower Bound} \\
	\mbox{GPR} & \mbox{Gaussian Process Regression} \\
	\mbox{GPCDE} & \mbox{Gaussian Process Conditional Density Estimation} \\
	\mbox{HNN} & \mbox{Heteroscedastic Neural Network} \\
	\mbox{\textit{i.i.d.}} & \mbox{Independent and Identically Distributed} \\
	\mbox{IWVI} & \mbox{Importance Weighted Variational Inference} \\
	\mbox{KL} & \mbox{Kullback-Leibler} \\
	\mbox{MCMC} & \mbox{Markov Chain Monte Carlo} \\
	\mbox{MDN} & \mbox{Mixture Density Network} \\
	\mbox{NLL} & \mbox{Negative Log Likelihood} \\
	\mbox{SGP} & \mbox{Scalable Gaussian Process} \\
	\mbox{SE} & \mbox{Squared Exponential} \\
	\mbox{SHGP} & \mbox{Scalable Heteroscedastic Gaussian Process} \\
	\mbox{SLGP} & \mbox{Scalable Latent Gaussian Process} \\
	\mbox{SMGP} & \mbox{Scalable Mixture of Gaussian Processes} \\
	\mbox{SVHGP} & \mbox{Stochastic Variational Heteroscedastic Gaussian Process} \\
	\mbox{VI} & \mbox{Variational Inference} \\
	\mbox{w.r.t.} & \mbox{With Respect To}
\end{array}
$

\subsection{Notation} 
$
\begin{array}{ll}
	\mathbf{a}^t, a^t_* & \mbox{Latent function values of }a^t\mbox{ }(0 \le t \le T)\mbox{ at }\mathbf{X}\mbox{ and }\mathbf{x}_*\mbox{ in SMGP} \\
	d_{\mbf{x}}, d_{\mbf{w}}, d_{\mbf{h}} & \mbox{Dimensionality of }\mbf{x}\mbox{, }\mbf{w}\mbox{ and }\mbf{h} \\
    \mathbf{h} & \mbox{Output of stochastic encoder in SLGP} \\
	\mathbf{K}, \mathbf{k} & \mbox{Kernel matrix and vector} \\
	\mathcal{L} & \mbox{Evidence lower bound} \\
	M & \mbox{Number of inducing variables} \\
	\mathbf{m}, \mathbf{S} & \mbox{Mean and covariane for the Gaussian }q(\mathbf{u}) \\
	\mathbf{m}^w, \mathbf{S}^w & \mbox{Mean and covariane for the Gaussian }q(\mathbf{u}^w)\mbox{ in SHGP} \\
	\mathbf{m}^{a,t}, \mathbf{S}^{a,t} & \mbox{Mean and covariane for the Gaussian }q(\mathbf{u}^{a,t})\mbox{ in SMGP} \\
	N & \mbox{Number of training points} \\
	\mathbf{f}, f_* & \mbox{Latent function values of }f\mbox{ at }\mathbf{X}\mbox{ and }\mathbf{x}_* \\
	\mathbf{f}^t, f^t_* & \mbox{Latent function values of }f^t\mbox{ at }\mathbf{X}\mbox{ and }\mathbf{x}_*\mbox{ in SMGP} \\
	\mathbf{w}, w_* & \mbox{Modulation variables at }\mathbf{X}\mbox{ and }\mathbf{x}_* \\
	S & \mbox{Number of monte carlo samples in SMGP and SLGP} \\
	T & \mbox{Number of assignment GPs in SMGP} \\
	\bm{\mu}^f,\bm{\Sigma}^f & \mbox{Mean and covariance for the Gaussian }q(\mathbf{f}) \\
	\bm{\mu}^{f,t},\bm{\Sigma}^{f,t} & \mbox{Mean and covariance for the Gaussian }q(\mathbf{f}^t)\mbox{ in SMGP} \\
	\bm{\mu}^w,\bm{\Sigma}^w & \mbox{Mean and covariance for the Gaussian }q(\mathbf{w})\mbox{ in SHGP} \\
	\bm{\mu}^{a,t},\bm{\Sigma}^{a,t} & \mbox{Mean and covariance for the Gaussian }q(\mathbf{a}^t)\mbox{ in SMGP} \\
	\mathbf{X}, \mathbf{y} & \mbox{Training inputs and outputs} \\
	\mathbf{x}_*, y_* & \mbox{Test inputs and outputs} \\
	\mathbf{Z}, \mathbf{u} & \mbox{Inducing points and variables} \\
	\mathbf{u}^w & \mbox{Inducing variables for the latent variables }\mathbf{w}\mbox{ in SHGP}  \\
	\mathbf{u}^{a,t} & \mbox{Inducing variables for the latent variables }\mathbf{a}^t\mbox{ in SMGP}  \\
	\bm{\nu}^f,\bm{\nu}^w & \mbox{Vectors comprising the diagonal elements of }\mathbf{\Sigma}^f\mbox{ and }\mathbf{\Sigma}^w\mbox{ in SHGP} \\
	\mathcal{B} & \mbox{Subset of training data} \\
	\mathcal{D} & \mbox{Training data} \\
	\mathcal{M} & \mbox{Multinomial distribution} \\
	\mathcal{N} & \mbox{Gaussian distribution} \\
	\beta & \mbox{Balance parameter in SLGP} \\
	\epsilon & \mbox{Observation noise} \\
\end{array}
$

\section{Experimental configurations} \label{app_exp_detail}
\textbf{The UCI datasets.} We elaborate the model configurations on the eight UCI datasets as below. First, we preprocess the data standardization over each dimension to have zero mean and unit variance. In addition, we shuffle and split the data to choose 90\% for training and the rest 10\% for testing.

Second, the modulated GPs employ the SE kernel with the length-scales initialized as $1.0$ and the signal variance initialized as $1.0$. They adopt $M=100$ inducing variables, the positions of which are initialized by the $k$-means clustering technique. The GP competitors including SGP, SVHGP, DAGP and GPCDE also take the same configuration for the fair of comparison. It is notable that, since the SLGP has an augmented input space $\mathbb{R}^{d_{\mbf{x}} + d_{\mbf{z}}}$, we draw samples from normal distribution to act as inducing points for the extra dimensions. Besides, the SLGP performs grid search of the balance parameter $\beta$ from the candidate set $\{1.0, 0.5, 0.1, 0.01\}$, uses a single latent input (i.e., $d_{\mbf{w}} = 1$), and has the latent dimension of $d_{\mbf{h}} = d_{\mbf{w}} + d_{\mbf{x}}$ without otherwise indicated. In addition, both the SMGP and SLGP use $S=10$ Monte Carlo samples to estimate their ELBOs~\eqref{eq_elbo_mgp} and~\eqref{eq_elbo_lgp_beta_hybrid}. Note that the SHGP model has no need due to the analytical ELBO~\eqref{eq_elbo_hgp}.

Third, as for the MLPs for $p(\mbf{w})$, $q(\mbf{w})$ and $q(\mbf{h}|\mbf{w})$ in SLGP, they take the fully-connected (FC) neural networks with three hidden layers, each of which has 100 units and applies the ReLU activation. Besides, the parameter $\nu_0$ is initialized as $0.01$ for the MLPs of encoder in~\eqref{eq_q(h|w)_mlp}. It is notable that the experiments do not use additional fine-tuning tricks, e.g., the scheduled learning rate, the regularized weights, or the dropout technique, for MLPs. The NN counterparts including HNN, MDN and CVAE also take the same architecture, with the only difference being that the last layer is not GP.

Forth, the training process adopts the Adam solver~\cite{kingma2014adam} with the batch size of 512, the learning rate of $0.005$, and the maximum number of iterations of 20000. 

Finally, as for model evaluation on the test set, we sample 200 points from the predictive distribution at each test point, and then employ the kernel density estimator from the \texttt{scikit-learn} package~\cite{scikit-learn} to estimate the density as well as the negative log likelihood. Note that the bandwidth parameter for the kernel density estimator is estimated by Silverman's rule.

\textbf{The large-scale \texttt{nytaxi} dataset.} The three scalable modulated GPs take almost the same configuration to that on the UCI datasets. The differences are that they herein adopt $M=500$ inducing variables and run the Adam optimizer with the batch size of 1024 and the maximum number of iterations of 50000; besides, the SLGP employs a small balance parameter $\beta=0.01$ on this large dataset to enhance extracting the patterns under the massive data; finally, the MLPs in SLGP and the NN models take the hidden architecture of ``FC(256)-Relu-FC(128)-Relu-FC(64)-Relu-FC(32)''.

\textbf{The $\mathtt{mnist}$ image dataset.} For this 10-input-784-output image dataset, we randomly choose 63000 images for training and the rest 7000 images for testing. The SLGP employs $M=100$ inducing points and the small balance parameter $\beta=0.01$, and runs the Adam optimizer over 50000 iterations with the batch size of 64. As for the MLPs in SLGP and CVAE, they take the hidden architecture of ``FC(512)-Relu-FC(512)-Relu-FC(512)''.
	
\end{appendices}

\section*{References}
%\footnotesize
\bibliography{ModulatedGP_ref}

\end{document}